%% file: acl_latex.tex
\PassOptionsToPackage{table}{xcolor}
\documentclass[11pt]{article}

\usepackage[final]{acl}
\usepackage{times}
\usepackage{latexsym}

\usepackage[T1]{fontenc}

\usepackage[utf8]{inputenc}

\usepackage{microtype}

\usepackage{inconsolata}

\usepackage{graphicx}

\usepackage{tcolorbox}
\tcbuselibrary{breakable, skins}
\usepackage{booktabs}
\usepackage{tabularx}
\usepackage{array}
\usepackage{multirow}
\usepackage{makecell}
\usepackage{bigdelim}
\usepackage{rotating}
\usepackage{pdflscape}
\usepackage{multirow}
\usepackage[table]{xcolor}
\usepackage{booktabs}
\definecolor{narrviolet}{RGB}{120, 60, 187}
\definecolor{funcorange}{RGB}{43, 105, 26}   %
\definecolor{identblue}{RGB}{37, 150, 190}
\definecolor{stereored}{RGB}{139, 0, 0}
\usepackage[table]{xcolor}
\usepackage[normalem]{ulem}
\usepackage{xcolor}

\usepackage{soul}
\definecolor{revyellow}{RGB}{255,242,150}

\usepackage{listings}                                                                           \lstdefinestyle{prompt}{
  basicstyle=\ttfamily\small,
  breaklines=true,            %
  breakindent=0pt,
  breakautoindent=true,
  frame=none,                 %
  backgroundcolor=\color{white},
  xleftmargin=1em,            %
  xrightmargin=1em,           %
  aboveskip=1em,
  belowskip=1em,
  columns=fullflexible,
  keepspaces=true,
}
\usepackage{amsmath}
\usepackage{url}

\input{commands}

\newcolumntype{Y}{>{\raggedright\arraybackslash}X}
\definecolor{stereored}{RGB}{139,0,0}        %
\definecolor{softred}{RGB}{180, 40, 40}
\definecolor{narrvioletbg}{RGB}{235, 225, 250}
\definecolor{audorangebg}{RGB}{225, 240, 220}
\definecolor{identbluebg}{RGB}{220, 240, 248}
\definecolor{modeviolet}{RGB}{120, 60, 187}   
\definecolor{funcorange}{RGB}{43, 105, 26}
\definecolor{identblue}{RGB}{37, 150, 190}   
\definecolor{softred}{RGB}{180, 40, 40}

\newcommand{\narrhl}[1]{\colorbox{narrvioletbg}{\strut #1}}
\newcommand{\audhl}[1]{\colorbox{audorangebg}{\strut #1}}
\newcommand{\identhl}[1]{\colorbox{identbluebg}{\strut #1}}

\title{
How Do People Challenge Racial Stereotypes Online? \\
Counter-Story Detection Across Reddit Communities \\

\vspace{0.1cm}
\small{\textit{\textcolor{softred}{Warning: content in this paper may be upsetting or offensive.}}}
}

\newcommand{\affsym}[1]{\textsuperscript{\ensuremath{#1}}}

\author{
  \textbf{Uma Sushmitha Gunturi}\affsym{\spadesuit} \quad
  \textbf{Jimin Mun}\affsym{\clubsuit} \quad
  \textbf{Maarten Sap}\affsym{\clubsuit} \quad
  \textbf{Maria Antoniak}\affsym{\heartsuit} \\[0.6em]
  \small
  \affsym{\spadesuit}IBM \quad 
  \affsym{\clubsuit}Carnegie Mellon University \quad
  \affsym{\heartsuit}University of Colorado Boulder \quad
}

\begin{document}
\maketitle
\begin{abstract}
\textit{Counter-storytelling} is a powerful mechanism people use to challenge dominant narratives. Unlike other forms of counterspeech that have been widely studied in computational social science, counter-storytelling has largely been overlooked. Counter-stories are
difficult to detect automatically; they are \textit{relational} (defined with respect to expressions of racial stereotypes) and \textit{structurally diverse} (drawing on stories that describe lived experiences, witnessed events, exemplars, and hypotheticals). We introduce a first framework for detecting and characterizing counter-storytelling against racial stereotypes at scale. This includes (1) a three-dimensional taxonomy grounded in narratology and Critical Race Theory and (2) a multi-stage pipeline that identifies relational pairs of stereotypes and counter-stories in noisy Reddit discourse. Using this pipeline, we annotate 25,549 Reddit posts across 615 communities and identify 1,312 counter-stories. Our analysis shows that speaker identity and post context shape how counter-stories are told. For example, in-group writers favor first-person testimony, often adopting the role of \emph{self-reflective insiders}. Our work shows how computational methods can scale qualitative approaches to identify and characterize counter-storytelling as a contextual narrative practice, with implications for content moderation, narratology, and racial discourse analysis.\footnote{Our pipeline code, race lexicon, stereotype carrier sentences, and annotation codebook are available at \url{https://github.com/UmaGunturi/counter_story_detection}}
\end{abstract}

\section{Introduction}
\label{sec:intro}

\begin{figure}[!t]

\definecolor{narrviolet}{RGB}{120, 60, 187}
\definecolor{audorange}{RGB}{43, 105, 26}
\definecolor{identblue}{RGB}{37, 150, 190}

\centering
\makebox[\columnwidth]{\includegraphics[width=1.05\columnwidth]{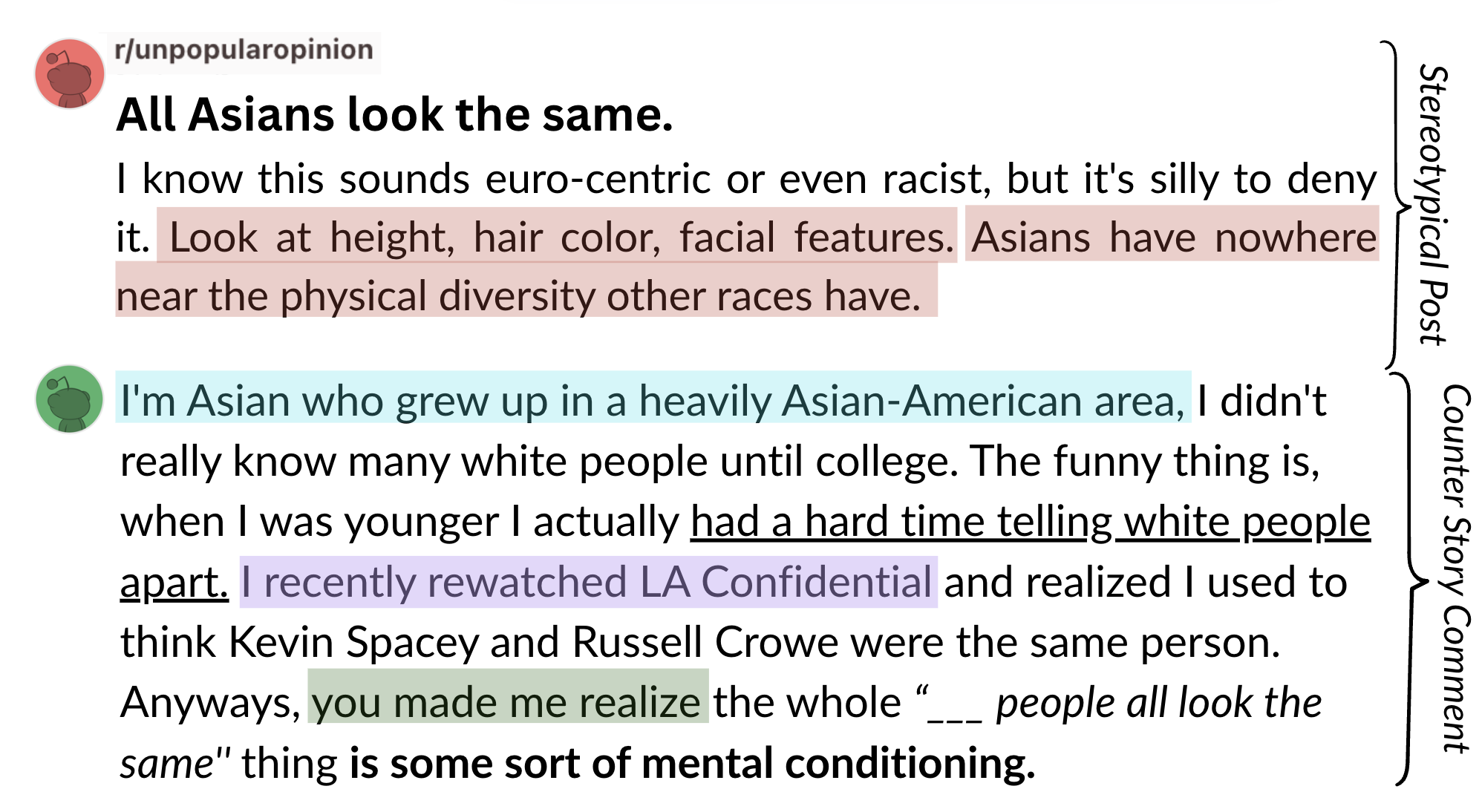}}
\caption{A motivating example of a counter-story responding to a stereotype-containing Reddit post (highlighted in red) on \texttt{r/unpopularopinion}. The example highlights the difficulty of identifying counter-storytelling: the narrator establishes a \textcolor{narrviolet}{\textbf{narrative position}} by recounting firsthand experience, asserts in-group \textcolor{identblue}{\textbf{identity}}, and directly addresses the original poster (\textcolor{audorange}{\textbf{audience}}). They also \protect\uline{invert} the racial premise of the stereotype and \protect\textbf{reframe} the stereotypical perception as mental conditioning rather than a fact about the group.}
\label{fig:example_cn_complex}
\end{figure}

\textit{Counter-storytelling} is the use of personal narratives to challenge dominant racial stereotypes---a practice that Critical Race Theory identifies as central to how marginalized communities resist racial oppression \cite{7bc14474-60ce-3264-9056-e8e63bc62c47, doi:10.1177/107780040200800103, Miles27052019, merriweather2006can}. 
For example, in Fig.~\ref{fig:example_cn_complex}, an Asian user responds to a stereotypical post claiming \textit{``all Asians look the same''.} They tell a counter-story that elegantly draws on lived experience, asserts in-group identity, and inverts the stereotype's premise. These stories can be more persuasive than factual arguments alone \citep{doi:10.1073/pnas.2008389118,Broockman2016DurablyRT, Lee2023StopAsianHateOT} and can foster inclusive attitudes \citep{Kalla2020ReducingEA}. 

Despite the power of these stories, most prior work in computational social science (CSS) has focused only on the broader discourse category of 
counter-speech, i.e., general responses against hate speech~\citep{bonaldi-etal-2024-nlp, fanton-etal-2021-human, mun-etal-2023-beyond}. 

But without a systematic understanding of counter-stories (who tells them, to whom, and from what position), we risk misinterpreting their social function. For example, moderation systems that cannot incorporate such context may flag counter-stories as toxic while failing to recognize the structural harms such stories contest \cite{Fraser2016RethinkingTP, Yang2025HowSR, Lee2024PeopleWS, Gunturi2024LinguisticallyDA, gligoric-etal-2024-nlp}.

In this work, we computationally examine counter-storytelling as it unfolds at scale across hundreds of Reddit communities. We formalize \textit{counter-story detection} as a \emph{relational} narrative task in a multi-step pipeline (App.~Fig.~\ref{fig:pipeline}) that (1) identifies a stereotype-containing post, (2) identifies a story-containing response, and (3) determines whether the personal narrative challenges the stereotype.

Using this pipeline, we construct a large-scale dataset of counter-stories against racial stereotypes, comprising 25,549 Reddit posts (of which 3,600 are stereotypes) and 6,538 story-containing comments (of which 1,312 are counter-stories) across 615 unique subreddits, with each counter-story labeled along a three-dimensional taxonomy of \emph{Narrative Position}, \emph{Audience}, and \emph{Identity} grounded in narratology and Critical Race Theory (\S\ref{sec:taxonomy}). This dataset enables empirical investigation, at a scale prior resources could not support.

Using this dataset, we analyze the associations between the different narrative features, and we examine how community context and stereotype type are associated with counter-stories.
We also perform a mixed-methods analysis to identify seven rhetorical roles, such as \emph{outgroup allies}, \emph{self-reflective insiders}, and \emph{ingroup mentors}, by surfacing common feature combinations across our three taxonomy dimensions. 
Our findings demonstrate the potential of computational methods to advance understanding of narrative resistance online.

In summary, our contributions include:
\begin{itemize}
\item a three-dimensional taxonomy of Narrative Position, Audience, and Identity, grounded in narratology and Critical Race Theory (\S\ref{sec:taxonomy}),
\item a multi-stage pipeline that adapts and extends prior stereotype-detection and story-identification methods for noisy, large-scale online discourse, enabling counter-story annotation at a scale that human labeling cannot reach (\S\ref{sec:data}),
\item a large-scale annotated dataset of 3,600 stereotypical posts and 1,312 counter-story comments across 209 subreddits, validated by human annotators (\S\ref{sec:data}),
\item and a comprehensive analysis of online counter-storytelling patterns, with implications for content moderation and platform design (\S\ref{sec:results}).
\end{itemize}

\section{Related Work}
\label{sec:related}

\paragraph{Counter-stories in CRT and beyond.}
Counter-storytelling has been studied across many disciplines as a way marginalized communities contest dominant accounts of social life.
These have included \textit{critical race theory} \citep{7bc14474-60ce-3264-9056-e8e63bc62c47,
doi:10.1177/107780040200800103, 10.1145/3313831.3376392, field-etal-2021-survey}, \textit{sociology} \citep{b1ff7e54-513b-35d3-a123-dacf9115db50, 10.1108/EDI-12-2022-0361}, \textit{literary theory} \citep{bal2009narratology, chatman1978story}, and \textit{public-sphere theory} on counterpublics \citep{Fraser2016RethinkingTP, warner2000counterpublics}. Within CRT, \citet{doi:10.1177/107780040200800103} describe ``majoritarian stories''---dominant narratives that frame assumptions about marginalized groups as natural or self-evident. Race scholarship further illustrates how such dominant narratives are constructed, defining marginalized groups in relation to dominant groups and institutions \citep{hegemony2015, long2014asianlatinx, Bonilla-Silva-2014, 10.1145/3351095.3372826}.
For example, scholarship has examined how racialized crime statistics constructed Blackness against a presumed-innocent white majority \citep{BannerHaley2011TheCO}, how the ``model minority'' stereotype positioned Asian Americans in relation to Black civil rights claims \citep{Li2015EllenDW}, and how representations of Latino immigrants cast them as a demographic threat to the nation \citep{NezMchiri2012TheLT}. We draw on this scholarship (App.~Table~\ref{tab:literature}) to identify the recurring racial stereotypes through which these dominant narratives persist in online discourse today.

Counter-stories are a powerful mechanism to respond to such narratives by centering the lived experiences of people of color \citep{Miles27052019, doi:10.1177/107780040200800103, merriweather2006can}. Their function is therefore relational: it depends on the dominant narrative they contest, the social position from which they are told \citep{7bc14474-60ce-3264-9056-e8e63bc62c47}, and the audience they address \citep{Fraser2016RethinkingTP, warner2000counterpublics}. We use this same structure when constructing our annotation framework.

More broadly, prior empirical work has shown that personal narratives can shape how audiences interpret contested social issues, increasing respect across moral and political divides \citep{Broockman2016DurablyRT, Kalla2020ReducingEA}. Yet counter-storytelling has never been studied at scale. We offer the first computational account of how counter-storytelling unfolds in online communities.

\paragraph{Counterspeech and counter narratives.}

Existing resources collect expert or crowd-authored counter-responses to hate speech, including CONAN \citep{Chung2019CONANC}, and related datasets spanning multiple languages and target communities \citep{Chung2019CONANC, fanton-etal-2021-human, bonaldi-etal-2024-nlp, yu-etal-2023-fine}. 

Counterspeech can include various rhetorical strategies, such as fact-correction, denouncing, humor, empathy, and perspective-taking \citep{Chung2021TowardsKC, mun-etal-2023-beyond}. 
These taxonomies capture argumentative and interpersonal strategies, but do not center counter-stories as a strategy or model narrative features.

Recent work has explored generating counter-narratives and evaluating them against expert-authored references \citep{Qian2019ABD, Tekirolu2020GeneratingCN, Saha2022CounterGeDiAC, Chung2021MultilingualCN, Furman2023AnIE, Mun2024CounterspeakersPU}. Rather than generating synthetic counter-stories, we examine how real people use counter-storytelling in their everyday lives online.

\section{A Counter-Stories Framework}
\label{sec:taxonomy}

Counter-stories can vary widely in who tells them, how narrators position themselves in relation to the events they describe, and whom the story is addressed to. To characterize this variation systematically, we develop a three-dimensional taxonomy capturing \emph{Narrative Position}, \emph{Identity}, and \emph{Audience} grounded in narratology and Critical Race Theory. Through iterative open coding of more than 300 counter-stories, we define preliminary dimensions (App. Fig.~\ref{fig:taxonomy}). 

\smallskip
\noindent
\narrhl{\textbf{Narrative Position}} captures the narrator's relationship to the story being told: whether the narrator stands outside the story world, appears within it, or recounts their own experience as the protagonist of the story \citep{Genette1980NarrativeD, chatman1978story, Fludernik2009AnIT}.

\smallskip\noindent
\textbf{Autodiegetic} narrators are simultaneously narrator and protagonist. They recount firsthand experiences of discrimination or resistance in the first person, usually as members of the victim group. 

\smallskip\noindent
\textbf{Homodiegetic} narrators are present in the story as witnesses rather than protagonists. This includes members of the victim group witnessing discrimination of others in their group, members of the dominant group recounting positive experiences with the victim group, and third-party observers witnessing interactions between groups.

\smallskip\noindent
\textbf{Heterodiegetic} narrators tell stories they were not present for, including: public exemplars, such as invoking public figures and historical events; second-hand narratives, which relay stories heard through trusted sources; and fictional narratives, which reference books, movies, and other media.

\smallskip\noindent
\textbf{Metadiegetic} narrators tell a story within a story, where the speaker's narrative includes another person's (e.g., quoted or reported) story.

\smallskip\noindent
\textbf{Hypothetical} narrators construct counterfactual scenarios, inviting the audience to reason through an alternative framing.

\smallskip
\noindent
\audhl{\textbf{Audience}} captures the implied primary recipient of the counter-story, since the same story can function differently depending on whom it addresses \citep{Fraser2016RethinkingTP, warner2000counterpublics}.
Audience can include: direct rebuttal of the \textit{Original Poster}; challenging or educating the \textit{Dominant Group} who perpetuates the stereotype; offering solidarity, guidance, or critique to the \textit{Victim
Group}; addressing the broader \textit{General/Mixed Audience}; and leaving the primary audience \textit{Ambiguous}.
Note that a single comment can address multiple audiences.

\smallskip
\noindent
\identhl{\textbf{Identity}} captures the narrator's social perspective, following CRT's emphasis on positionality as constitutive of counter-storytelling \citep{doi:10.1177/107780040200800103},
who is speaking shapes what the story can claim and how it is likely to be received.
Identity includes explicit self-disclosures of \textit{race}, \textit{gender}, and \textit{connection} to the target group, separating narrators speaking as members of the stereotyped group from witnesses or allies.

\section{Data and Annotation Pipeline}
\label{sec:data}
We apply our framework to a large dataset of Reddit posts and comments drawn from many diverse communities. 
We use a combination of lexicons, hand annotation, and LLM-assisted prediction to first identify posts containing stereotypes and then identify comments replying to those posts that contain counter-stories (App.~Fig.~\ref{fig:pipeline} provides an overview of the full pipeline).

\begin{table}[t]
\centering
\small
\begin{tabular}{llrr}
\toprule
\textbf{Type} & \textbf{Label} & \textbf{\#} & \textbf{\%} \\
\midrule
\multirow{6}{*}{Post}
  & \colorbox{stereored!20}{Stereotypes}    & 3,600 & 14.1 \\
  & \quad Asian                                 & 1,198 & 33.3 \\
  & \quad Black                                 & 868 & 24.1 \\
  & \quad Hispanic                              & 71 & 2.0 \\
  & \quad Other                                 & 1,463 & 40.6 \\
  & \colorbox{gray!15}{Non-stereotypes}     & 21,949 & 85.9 \\
\addlinespace
\multirow{6}{*}{Comment}
  & \colorbox{orange!20}{Counter-stories}   & 1,312 & 20.1 \\
  & \quad Asian                                 & 484 & 36.9 \\
  & \quad Black                                 & 351 & 26.8 \\
  & \quad Hispanic                              & 34 & 2.6 \\
  & \quad Other                                 & 443 & 33.8 \\
  & \colorbox{gray!15}{Non-counter-stories} & 5,226 & 79.9 \\
\bottomrule
\end{tabular}
\caption{Overview of the LLM annotated dataset.}
\label{tab:dataset_stats}
\end{table}

\subsection{Data Collection and Filtering}

\paragraph{Data source.}
We construct our dataset from two Reddit sources. For discussions prior to 2020, we use ConvoKit~\cite{chang-etal-2020-convokit}, which contains $\sim$948K subreddits.
We also collect data via the Arctic Shift API\footnote{\href{https://github.com/ArthurHeitmann/arctic_shift}{https://github.com/ArthurHeitmann/arctic\_shift}} for conversations 2020--May 2026.

\smallskip
\noindent
\paragraph{Race lexicon.}
We construct a lexicon of racial identity terms (e.g., ``Black,'' ``Asian''), slurs, and coded language for three focal groups: \emph{Black}, \emph{Asian}, and \emph{Hispanic}. Lexicon construction and the full term list are in App. \S\ref{sec:race_lexicon_appendix}.

\smallskip
\noindent
\paragraph{Community curation.}
We filter the ConvoKit corpus from $\sim$948K subreddits to a focused
working set that is more likely to contain counter-storytelling.
(1) Subreddits with fewer than 1{,}000 users, NSFW spaces, and quarantined communities are dropped. Within the remaining communities, we retain only those with at least one post containing a race-lexicon term. This leaves approximately 37{,}000 candidate subreddits spanning over 20 million posts.
(2) We require each subreddit to contain at least 100 cumulative identity-keyword mentions, ensuring that race is a recurring rather than incidental topic.
(3) For each subreddit, we compute a \textit{match ratio}: the proportion of its posts (after the initial filtering step) that contain at least one lexicon term. We retain the top 1000 subreddits with match ratio $\geq 0.5$. This step filters communities where race-related discussion is concentrated.

\smallskip
\noindent
\paragraph{Post sampling and story detection.}
From these subreddits, we collect $\sim$150K lexicon-matched posts and retain only posts whose comment threads contain at least one story, applying StorySeeker~\cite{antoniak-etal-2024-people}, a classifier trained to detect stories on Reddit, to all top-level comments. We filter comments with <30 words and retain at most 500 story comments per post. This yields 90{,}061 posts with at least one story comment.

\begin{table*}[t]
\centering
\small
\setlength{\tabcolsep}{2.5pt}
\renewcommand{\arraystretch}{1.1}
\begin{tabular}{lrrrrrrrrrrr}
\toprule
& & & & \multicolumn{3}{c}{\textit{Narrative Position}} & \multicolumn{3}{c}{\textit{Audience}} & \multicolumn{2}{c}{\textit{Identity}} \\
\cmidrule(lr){5-7} \cmidrule(lr){8-10} \cmidrule(lr){11-12}
\textbf{Domain} & \textbf{\#Subs} & \textbf{\#CS} & \textbf{CS\%} & \makecell[c]{\textbf{Auto-}\\\textbf{diegetic}} & \makecell[c]{\textbf{Hetero-}\\\textbf{diegetic}} & \makecell[c]{\textbf{Homo-}\\\textbf{diegetic}} & \makecell[c]{\textbf{Original}\\\textbf{Poster}} & \makecell[c]{\textbf{General/}\\\textbf{Mixed}} & \makecell[c]{\textbf{Victim}\\\textbf{Group}} & \makecell[c]{\textbf{Target}\\\textbf{Group}} & \makecell[c]{\textbf{Dominant}\\\textbf{Group}} \\
\midrule
\textit{Culture/Identity} & 66 & 430 & 20.9\% & \cellcolor{narrviolet!50}68\% & \cellcolor{narrviolet!25}8\% & \cellcolor{narrviolet!25}12\% & \cellcolor{funcorange!50}98\% & \cellcolor{funcorange!25}86\% & \cellcolor{funcorange!80}35\% & \cellcolor{identblue!80}66\% & \cellcolor{identblue!10}5\% \\
\textit{Advice/Education/Work} & 38 & 300 & 20.3\% & \cellcolor{narrviolet!80}72\% & \cellcolor{narrviolet!25}9\% & \cellcolor{narrviolet!10}10\% & \cellcolor{funcorange!80}99\% & \cellcolor{funcorange!10}80\% & \cellcolor{funcorange!50}23\% & \cellcolor{identblue!50}33\% & \cellcolor{identblue!50}17\% \\
\textit{Debate/Opinion} & 7 & 200 & 36.4\% & \cellcolor{narrviolet!10}34\% & \cellcolor{narrviolet!50}30\% & \cellcolor{narrviolet!50}14\% & \cellcolor{funcorange!50}98\% & \cellcolor{funcorange!50}98\% & 0\% & \cellcolor{identblue!25}10\% & \cellcolor{identblue!50}14\% \\
\textit{General/Ask} & 19 & 140 & 24.0\% & \cellcolor{narrviolet!25}54\% & \cellcolor{narrviolet!25}15\% & \cellcolor{narrviolet!80}19\% & \cellcolor{funcorange!80}100\% & \cellcolor{funcorange!50}99\% & \cellcolor{funcorange!25}3\% & \cellcolor{identblue!50}26\% & \cellcolor{identblue!80}31\% \\
\textit{Stories} & 13 & 93 & 12.3\% & \cellcolor{narrviolet!50}65\% & \cellcolor{narrviolet!10}8\% & \cellcolor{narrviolet!25}12\% & \cellcolor{funcorange!25}98\% & \cellcolor{funcorange!10}80\% & \cellcolor{funcorange!50}25\% & \cellcolor{identblue!50}25\% & \cellcolor{identblue!80}28\% \\
\textit{Geography} & 28 & 53 & 22.2\% & \cellcolor{narrviolet!25}51\% & \cellcolor{narrviolet!50}15\% & \cellcolor{narrviolet!50}15\% & \cellcolor{funcorange!50}98\% & \cellcolor{funcorange!25}94\% & \cellcolor{funcorange!25}6\% & \cellcolor{identblue!25}13\% & \cellcolor{identblue!25}13\% \\
\textit{Humor} & 8 & 39 & 13.6\% & \cellcolor{narrviolet!25}38\% & \cellcolor{narrviolet!80}33\% & \cellcolor{narrviolet!80}18\% & \cellcolor{funcorange!10}87\% & \cellcolor{funcorange!80}100\% & 0\% & \cellcolor{identblue!10}3\% & \cellcolor{identblue!25}5\% \\
\textit{Hobbies} & 14 & 24 & 16.6\% & \cellcolor{narrviolet!80}71\% & \cellcolor{narrviolet!50}17\% & \cellcolor{narrviolet!10}8\% & \cellcolor{funcorange!25}96\% & \cellcolor{funcorange!80}100\% & 0\% & \cellcolor{identblue!10}4\% & \cellcolor{identblue!10}4\% \\
\textit{Misc} & 8 & 15 & 4.9\% & \cellcolor{narrviolet!80}73\% & \cellcolor{narrviolet!10}7\% & \cellcolor{narrviolet!25}13\% & \cellcolor{funcorange!25}93\% & \cellcolor{funcorange!25}87\% & \cellcolor{funcorange!80}33\% & \cellcolor{identblue!80}33\% & \cellcolor{identblue!25}7\% \\
\textit{Politics} & 5 & 12 & 13.8\% & \cellcolor{narrviolet!10}33\% & \cellcolor{narrviolet!80}42\% & \cellcolor{narrviolet!50}17\% & \cellcolor{funcorange!10}92\% & \cellcolor{funcorange!80}100\% & 0\% & \cellcolor{identblue!25}8\% & 0\% \\
\textit{Beauty/Fashion} & 3 & 6 & 14.0\% & \cellcolor{narrviolet!50}67\% & 0\% & 0\% & \cellcolor{funcorange!80}100\% & \cellcolor{funcorange!80}100\% & 0\% & \cellcolor{identblue!80}33\% & \cellcolor{identblue!50}17\% \\
\midrule
\textbf{Total} & \textbf{209} & \textbf{1312} & \textbf{20.1\%} & 60\% & 14\% & 13\% & 98\% & 88\% & 19\% & 36\% & 14\% \\
\bottomrule
\end{tabular}
\caption{Counter-stories grouped by community domain, ordered by counter-story count. \textbf{\#Subs}: subreddits with $\geq 1$ counter-story. \textbf{\#CS}: total counter-stories. \textbf{CS\%}: share of comments labeled counter-stories. Audience is multi-label, so columns may sum $>$100\%. The remaining narrative positions (hypothetical, metadiegetic) are reported in App. Table~\ref{tab:domain_rollup_appendix}. Speaker's gender and race breakdowns are reported separately in Table~\ref{tab:gender_position} and Table~\ref{tab:race_position}.}
\label{tab:domain_rollup_combined}
\end{table*}

\subsection{Identifying Posts with Racial Stereotypes}
\label{subsection-labeling-stereotypes}

\paragraph{Concept-carrier semantic filter.}
Building on \citet{mendelsohn2025peoplefloods, sap-etal-2020-social}, we operationalize racial stereotypes as a set of \emph{concepts} across three target groups: Black (e.g., \textit{crime}), Asian (e.g., \textit{model minority}), and Hispanic (e.g., \textit{immigration}). We construct a concept inventory  
by defining 66 concepts across the three target groups. For each concept, we construct 10 carrier sentences (5 implicit and 5 explicit), 
which are written to express the stereotype with minimal semantic content (e.g., implicit: ``They always seem to avoid work''; explicit: ``They are lazy''). The full concept inventory, its sources from race scholarship, and example carriers per concept are in App.~\S\ref{sec:concepts_stereotpe_carrier_sentences} (Tables~\ref{tab:literature}--\ref{tab:carriers_hispanic}) and App.~\S\ref{appendix:intersectional_stereotypes}. For each concept, we compute centroid embeddings as the mean of its carriers (separately for its implicit and explicit variants) using Sentence-BERT (\texttt{all-MiniLM-L6-v2}; \citealt{sentence-bert-2019}). For each post, we extract the first sentence containing a race-lexicon keyword (matched at the token level) and compute its cosine similarity to every race-restricted concept centroid. 
We assign each post the concept with the highest score. Posts with  $\max(\text{similarity}) > 0.30$ advance to the next stage.\footnote{The $0.30$ threshold was selected by inspecting samples at threshold values  between $0.20$ and $0.50$; lower thresholds contained too many off-topic posts, higher thresholds excluded subtler stereotype posts.}

\begin{table}[htpb]
\centering\small
\setlength{\tabcolsep}{6pt}
\renewcommand{\arraystretch}{1.15}
\begin{tabular}{l|cc|cc}
\toprule
& \multicolumn{2}{c|}{\textbf{Stereotype}}
& \multicolumn{2}{c}{\textbf{Counter-Story}} \\
\textbf{Annotator}
& \textbf{Agr.\%} & \textbf{$\kappa$}
& \textbf{Agr.\%} & \textbf{$\kappa$} \\
\midrule
\textit{Human} & 86 & 0.73 & 82 & 0.64 \\
\midrule
\cellcolor{gray!18}Sonnet 4.6
              & \cellcolor{gray!18}\textbf{88}
              & \cellcolor{gray!18}\textbf{0.76}
              & \cellcolor{gray!18}\textbf{86}
              & \cellcolor{gray!18}\textbf{0.72} \\
\bottomrule
\end{tabular}
\caption{Inter-annotator agreement (Fleiss' $\kappa$) across three annotators and model agreement with the human majority vote (Cohen's $\kappa$), on 50 stereotype posts and 100 counter-story comments.}
\label{tab:human_validation}
\end{table}

\medskip\noindent
\textbf{LLM assisted filtering.} 
We label each of the 26K
candidate posts (that passed the concept-carrier stereotype filter) with Claude Sonnet 4.6 using the few-shot prompt shown in App.~\S\ref{app:stereotype_prompt}. The model classifies whether each post expresses a racial stereotype,
and if so, labels the victim group (Black, Asian, Hispanic, Other), stereotype type (e.g., Criminality),
a certainty score (1–10) and a rationale. A total of 3.6K (14.1\%) are labeled as expressing stereotypes (see  Table~\ref{tab:dataset_stats} for detailed distribution stats). 
Note that the lexicon is used only to retrieve candidate posts via keyword matching; the target group and stereotype type are also assigned independently by the model from the full post content.

\subsection{Labeling Counter-Stories}
For the 3{,}600 stereotype-containing posts, we extract 6{,}538 story comments in total and prompt Claude Sonnet 4.6 with the parent post, the comment, and our codebook (\S\ref{sec:taxonomy}) to (i) decide whether the comment functions as a counter-story and (ii) annotate it along \emph{Narrative Position}, \emph{Audience}, and \emph{Identity}. 1{,}312 comments (20.1\%) are labeled as counter-stories, spanning 209 unique subreddits. We provide the full codebook in App.~\S\ref{sec:codebook} and the prompt in App.~\S\ref{sec:appendix_prompts_counter_narratives}.

\subsection{Human Validation}
\label{sec:validation}
To assess how closely the model's labels align with human judgment, we conducted
iterative human validation for both stereotype and counter-story labeling. Across multiple rounds, three human annotators independently annotated 50 stereotype examples and 100 counter-story examples, achieving substantial inter-annotator agreement ($\kappa=0.73$ for stereotypes, $\kappa=0.64$ for counter-stories) with disagreements used to refine the codebook and annotation guidelines. We then compared candidate LLM annotators against human labels; Claude Sonnet~4.6 showed the strongest agreement (Table~\ref{tab:human_validation} and App.~Table~\ref{tab:backbone_comparison}). We further report precision/recall/F1 and find that the model's certainty scores are well calibrated with its accuracy (App.~Tables~\ref{tab:prf},~\ref{tab:calibration}). A false-negative audit of $\sim$100 randomly sampled story comments found only $\sim$2\% missed counter-stories, suggesting that the counter-stories missed by the pipeline are relatively uncommon (App.~\ref{appendix:fn_audit}).

\section{Analysis}
\label{sec:results}

Given our annotated dataset of stereotype posts and counter-story comments spanning many diverse communities, we explore a series of questions that benefit from its scale.

\subsection{Narrative Dimensions of Online Counter-stories}
\label{sec:rq1}
To understand the narrative diversity of counter-stories told online, we examine the distribution of the three narrative dimensions: narrative position, audience, and identity. We discuss these results as shown in Table~\ref{tab:domain_rollup_combined} (App.~Table~\ref{tab:subreddit_categories} lists example subreddits per domain).

\smallskip
\noindent
\textbf{Counter-stories are most commonly told by auto-diegetic narrators.} We observe that 60\% of counter-stories found in our dataset were told using auto-diegetic narrative positions, followed by 14\% heterodiegetic  and 13\% homodiegetic. The distribution differed based on community domains with Misc (73\%), Advice/Education/Work (72\%), Hobbies (71\%), and Culture/Identity (68\%) having the highest rate of autodiegetic narrators. Interestingly, we observe higher heterodiegetic narrators than autodiegetic narrators in Politics, 42\% and 33\% respectively, and similar rates of hetero and autodiegetic narrators in Humor, 33\% and 38\%. The prevalence of autodiegetic narrators suggests that many counter-stories told online come from first-hand personal experiences. Moreover, the difference based on domain implies that narrative positions are contextual and community dependent. The elevated counter-story rate in Debate/Opinion (36.4\%) is concentrated in \textit{r/changemyview} (70.5\%), a persuasion-oriented community \citep{antoniak-etal-2024-people} (App.~Table~\ref{tab:cmv_breakdown}).

\smallskip
\noindent
\textbf{Original posters are the most frequent audience.}
As shown in Table~\ref{tab:domain_rollup_combined}, the most frequent audience of counter-stories are the original posters who expressed the stereotype (98\%), followed by a general/mixed audience (88\%), and the victim group (19\%). This pattern is statistically robust (McNemar's test, $p<0.001$) for domains such as Culture/Identity, Advice/Education/Work, and Stories. Moreover, both original posters and general/mixed audience have high frequencies, suggesting that counter-storytellers often address both audiences in one story.

\smallskip
\noindent
\textbf{Counter-storytellers often belong to the target group.}
While most counter-storytellers do not disclose their group membership (53\%), those who do are more often members of the target group (36\%) than the dominant group (14\%). Interestingly, we see a high percentage of in-group counter-storytelling in the Culture/Identity domain (66\%), suggesting more in-group sharing and discussions. We hypothesize that in-group membership especially adds credibility to subjective experiences in this domain.\looseness=-1

\begin{table}[t]
\centering
\small
\begin{tabular}{lrrr}
\toprule
\textbf{Stereotype Type} & \textbf{Asian} & \textbf{Black} & \textbf{Other} \\
\midrule
Cultural       & 74.3\% & 52.6\% & 76.5\% \\
Appearance     & 10.6\% & 12.4\% &  5.7\% \\
Criminality    &  2.9\% & 13.6\% &  6.6\% \\
Dehumanization &  4.7\% &  9.4\% &  3.9\% \\
Other          &  3.4\% &  6.8\% &  2.9\% \\
Intelligence   &  3.2\% &  3.1\% &  1.8\% \\
Economic       &  0.9\% &  2.0\% &  2.6\% \\
\midrule
$n$ posts & 1{,}198 & 868 & 1{,}463 \\
\bottomrule
\end{tabular}
\caption{Distribution of stereotype types by target group (column-wise within each group). }
\label{tab:stereotype_dist_main}
\end{table}

\subsection{Influence of Context on Narrative Axes}
\label{sec:rq2}
\begin{figure}[t]
\includegraphics[width=1.00\columnwidth]{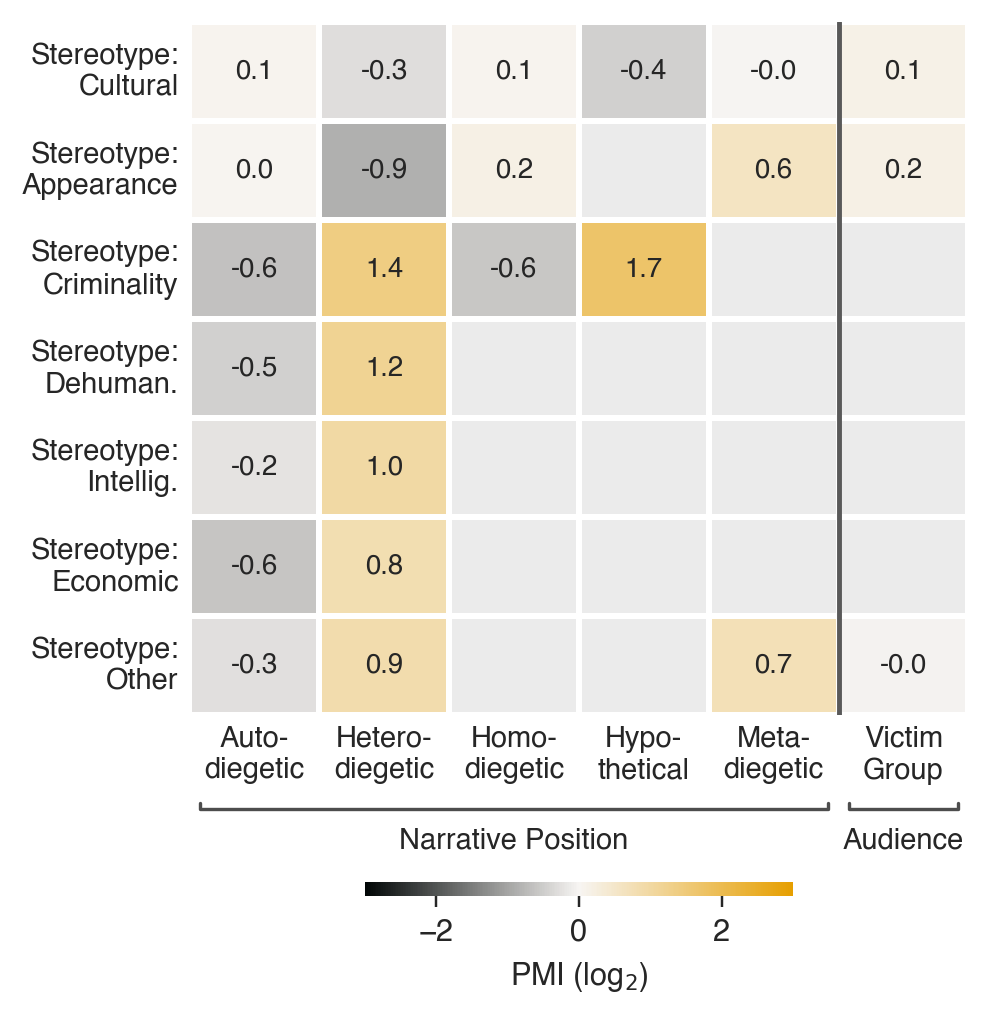}
\caption{Pointwise mutual information (PMI, $\log_2$) between stereotype context and narrative dimensions across all the 1{,}312 counter-stories. Orange indicates feature pairs that co-occur more than chance; blank cells indicate no co-occurrence. Gray cells denote pairs with fewer than five co-occurrences ($n<5$).}
\label{fig:pmi_stereotype_full}
\end{figure}

\begin{figure}[t]
\includegraphics[width=1.00\columnwidth]{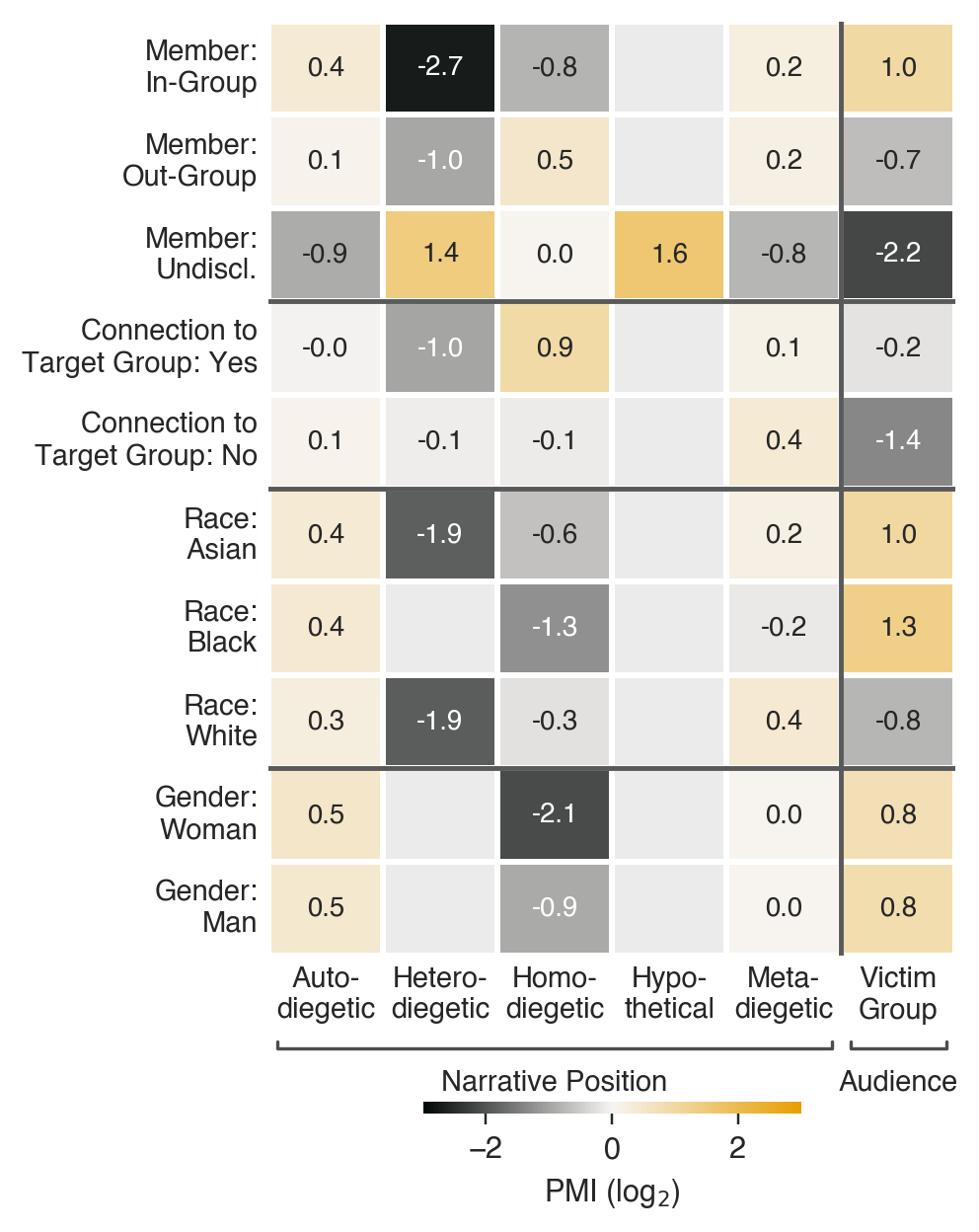}
\caption{Pointwise mutual information (PMI, $\log_2$) between speaker context and narrative dimensions across all the 1{,}312 counter-stories. Orange indicates feature pairs that co-occur more than chance; blank cells indicate no co-occurrence. Gray cells denote pairs with fewer than five co-occurrences ($n<5$). Additional feature pairs are reported in App.~\S\ref{appendix:additional_results} (Fig.~\ref{fig:pmi_appendix}).}
\label{fig:pmi_identity_full}
\end{figure}

To understand the relational aspects of counter-storytelling, we analyze the contexts in which counter-storytelling occurs. First, we observe the \textit{post context}: the stereotype type and the target group of the stereotype. Second, we consider the \textit{speaker context}: how the writer's disclosed relationship to the target group, race, and gender are associated with narrative dimensions. To surface these associations, we compute pointwise mutual information (PMI) scores between feature pairs across the 1{,}312 counter-stories (Figures~\ref{fig:pmi_stereotype_full} and~\ref{fig:pmi_identity_full}).\looseness=-1

\smallskip\noindent
\paragraph{Types of stereotype vary across racial identities.}
Stereotypes expressed in posts span seven categories across the three racial groups (see Table~\ref{tab:stereotype_dist_main}). Cultural stereotypes dominate across all groups, but notably, criminality and dehumanization stereotypes are most prevalent in Black-targeted posts (13.6\%, 9.4\%).
By contrast, Asian-targeted posts are almost entirely cultural (74.3\%) with comparatively little criminality (2.9\%). See App.~\S\ref{appendix:additional_results} (Table~\ref{tab:stereotype_rollup_v2}) for stereotype distributions across all domains. We find that Hispanic-targeted content is sparsely represented in our data (71 stereotype posts and 34 counter-stories, $\sim$2\% of each category). We therefore restrict cross-group interpretation to the Black and Asian groups and report Hispanic statistics in App.~\S\ref{appendix:hispanic_content}.\looseness=-1

\smallskip\noindent
\paragraph{Stereotypes are associated with narrative form.}
Cultural and appearance stereotypes sit near chance across narrative positions (PMI range $-0.4$ to $0.2$, Fig.~\ref{fig:pmi_stereotype_full}) (raw proportions in App.~Table~\ref{tab:stereotype_position}), suggesting they are countered in many different forms. Criminality and dehumanization stereotypes, however, co-occur more than chance with heterodiegetic narration (PMI~$1.4$, $1.2$) and hypothetical narration (PMI~$1.7$, n=9). Counter-stories addressing criminality stereotypes frequently draw on public exemplars and documented cases (criminality $\times$ exemplar PMI~$1.8$; see App. Fig.~\ref{fig:pmi_appendix}). We also observe that the most stigmatizing stereotypes are less likely to be addressed \emph{to} the targeted group: while cultural and appearance stereotypes sit near chance with victim-group address (PMI~$0.1$, $0.2$), more stigmatizing stereotypes tend to orient toward the original poster or a general audience.\looseness=-1

\smallskip
\noindent
\paragraph{Speaker context is associated with narrative position.}
The writer's disclosed relationship to the stereotyped group is strongly associated with how they tell counter-stories (Fig.~\ref{fig:pmi_identity_full}). Narrators in the target group co-occur with addressing the victim group (PMI $1.0$).  On the other hand, in-group members rarely co-occur with heterodiegetic (PMI $-2.7$) narration. This suggests that in-group members tend to speak \emph{as} insiders \emph{to} insiders, grounding their counter-story in lived experience and addressing those who share it.\looseness=-1

Moreover, writers who report a personal connection without claiming membership tend to use homodiegetic witness narration (PMI $0.9$) and who disclose no relationship to the targeted group show the opposite pattern: they co-occur more with heterodiegetic (PMI $1.4$) and hypothetical narration (PMI $1.6$), and less with victim-group address (PMI $-2.2$). This extends to narrative subtype: writers who disclose no membership lean on public exemplars (PMI~$1.5$) and fictional references (PMI~$1.7$, App.~\S\ref{appendix:fiction_examples}) 
(in-group $\times$ exemplar PMI~$-2.6$; App. Fig.~\ref{fig:pmi_appendix}). Narrative-position distributions by writer identity and target group are shown in App.~Fig.~\ref{fig:position_breakdown}. \looseness=-1

Similarly, among writers who disclose their race, Black and Asian writers are associated more with victim-group address (PMI $1.3$, $1.0$), unlike White writers (PMI $-0.8$) who are more likely to address the OP or a general audience.

\paragraph{Context alone does not predict counter-stories.} We further examine whether \emph{context} features and \emph{speaker} features predict when a story-containing comment functions as a counter-story, by fitting logistic regression models over all story comments. Except in-group membership, no other feature emerges as a robust independent predictor, reinforcing that counter-storytelling may not be driven by any single factor but instead arise from \textit{combinations} of speaker position, audience, stereotype type, and community context (App.~\ref{appendix:regression}).\looseness=-1

\subsection{Rhetorical Roles in Counter-Storytelling}
\label{sec:rq3}
What kinds of rhetorical roles do counter-storytellers take on when challenging racial stereotypes? We build on prior work that clusters behavioral patterns of online community members \citep{Yang2019SeekersPW, Sun2019MultilevelMO, Mason2025RacismRA}, and represent each of the 1,312 counter-stories as an 11-variable feature vector spanning the three taxonomy dimensions: narrative position, identity, and audience. 
We fit categorical Latent Class Analysis (LCA) \citep{Collins2009LatentCA} models for $K = 2$ through $K = 12$ using the \texttt{stepmix} library \citep{Morin2023StepMixAP} and selected the model with the lowest BIC at $K = 8$. The first author then qualitatively interpreted the latent classes by reading the 20 most confidently assigned examples from each class and examining each class's label profile to understand the social and rhetorical role the commenter was taking on. Through an iterative process, we merged overlapping classes and assigned role names to the final classes. This yielded seven final counter-storytelling roles, shown in Table~\ref{tab:role_counts}. The two merged classes include one almost purely heterodiegetic (Class 4, n=138) and one containing heterodiegetic and metadiegetic stories (Class 7, n=83), both citing external evidence. We qualitatively inspected these classes and merged them into one single class \textit{``Example Provider''} (n= 221). The full BIC curve is in App.~Fig.~\protect\ref{fig:bic}.\looseness=-1

\begin{table}[t]
\centering
\small
\begin{tabular}{lrr}
\toprule
\textbf{Role} & \textbf{$n$} & \textbf{\%} \\
\midrule
Outgroup Ally           & 344 & 26.2 \\
Self-Reflective Insider & 276 & 21.0 \\
Example Provider        & 221 & 16.8 \\
Empathetic Reframer     & 207 & 15.8 \\
Ingroup Mentor          & 173 & 13.2 \\
Outgroup Reassurer      &  52 &  4.0 \\
Context Provider        &  39 &  3.0 \\
\midrule
\textbf{Total}          & \textbf{1{,}312} & \textbf{100.0} \\
\bottomrule
\end{tabular}
\caption{Distribution of counter-stories across the seven counter-storytelling roles identified by LCA. Total $N = 1{,}312$.}
\label{tab:role_counts}
\end{table}

\smallskip
\noindent
\textbf{Outgroup Allies} counter stereotypes from outside the targeted group by leveraging their dominant-group position to reject prejudice within their own communities (``\textit{As a white person, I've heard my relatives make these kinds of comments, and they're simply wrong}''), often using personal experience to contest the stereotype (``\textit{Some of the kindest, hardest-working people I know are immigrants...}'').

\smallskip
\noindent
\textbf{Self-Reflective Insiders} counter stereotypes from within the targeted group through introspective, often deeply personal stories about identity, family, appearance, relationships, and social expectations (``\textit{I have felt the pressures of the stereotype of the studious, perfect Asian kid but it never really seemed to bother me. I knew deep down that I never really fit into that [...]}''). These comments are frequently autodiegetic as insiders narrate how stereotypes are absorbed, resisted, and reinterpreted from within the group.

\smallskip
\noindent
\textbf{Ingroup Mentors} are members of the stereotyped group who respond to a fellow member's struggle by sharing their own parallel experience, then offering emotional support and concrete life advice. Rather than only refuting the stereotype, these counter-stories model survival, validation, and guidance for an in-group audience. Some examples include: shared experience (``\textit{my dad was the same}''), reassurance (``\textit{you deserve better}'', ``\textit{I'm here}''), practical advice (``\textit{speak to your doctor}'', ``\textit{seek CBT}'', ``\textit{my tip is\ldots}''), well-wishing (``\textit{I hope you're okay}'').

\smallskip
\noindent
\textbf{Empathetic Reframers} ask the original poster or broader audience to reinterpret a stereotyped situation with more care. They counter by shifting attention from racialized blame to alternative explanations, such as stress, grief, exhaustion, family dynamics, or situational constraint. For example, after an OP describes his Chinese partner and her family as ``\textit{more cold},'' one commenter recalls, ``\textit{I had an emergency c-section and my immediate family came only to help me\ldots\ With her being a new mom, sleep-deprived and breastfeeding, she's probably struggling. Give her some grace\ldots\ she's probably just scared.}''

\smallskip
\noindent
\textbf{Outgroup Reassurers} counter stereotypes from outside the targeted group by offering comfort and affirmation. They often speak directly to the stereotyped group and use their own experiences (e.g., dating, friendship, or family) to reassure stereotyped group. These comments are less about proving the stereotype wrong and more about reassuring the target that they are not alone. For example, one white commenter describes marrying an Indian partner despite family disapproval and tells the target: ``\textit{being Indian is not something to hate about yourself; find people who love you and do not make you feel ashamed}.''\looseness=-1

\smallskip
\noindent
\textbf{Context Providers} counter stereotypes by unpacking the broader
conditions behind the behavior being racialized, such as labor exploitation, migration constraints, institutional failure, economic precarity, or unequal access to resources. For example, in response to the stereotype that ``\textit{Indians take jobs unfairly},'' one commenter reframes: ``\textit{getting such jobs is not easy [...] my brother competed in an extremely selective admissions system, spent years studying long hours, and gave up much of regular school and social life to get there}.''\looseness=-1

\smallskip
\noindent
\textbf{Example Providers} counter stereotypes by offering concrete cases e.g., stories, jokes, film references, public examples (such as linking a Washington Post list of 100 racially miscast Hollywood roles), or relational anecdotes as evidence. These comments challenge generalizations by supplying counterexamples in different forms to contest the stereotype.\looseness=-1

\section{Discussion}
\label{sec:discussion}

\paragraph{Counter-storytelling is a relational practice.} 
Our findings confirm that counter-storytelling is fundamentally \textit{relational}, consistent with CRT's emphasis on positionality \cite{doi:10.1177/107780040200800103}. Despite modeling a broad set of contextual and speaker-level features, only in-group membership remains a robust predictor of counter-storytelling after multiple-comparison corrections (\S\ref{sec:rq2}). This pattern reinforces that whether a story functions as a counter-story depends on the interaction among speaker position, audience, and identity. Our latent role analysis further supports this interpretation, as the seven rhetorical roles emerge as recurring combinations of these dimensions. For example, \textit{Outgroup Allies} combine out-group speaker position, heterodiegetic narration, and dominant-group address. At the same time, this relationality presents a challenge for computationally detecting counter-stories. Models that treat counter-storytelling as a property of isolated comments, without the surrounding stereotype post or interactional context, may mischaracterize how counter-stories function in practice \cite{Gunturi2024LinguisticallyDA, 10.1145/3351095.3372826}. Indeed, during our annotation process for counter-stories, we found that including the original post was essential for annotators to label the replies accurately. This also emphasizes the importance of community involvement and engagement, as interpretation of counter-stories relies on contextual knowledge. More broadly, computational approaches to racial discourse should therefore treat interaction and context as fundamental to understanding how racial narratives are reproduced and contested.\looseness=-1

\paragraph{Implications for content moderation.}
Our findings highlight a challenge for moderation systems: counter-stories can resemble the harmful content they contest. The most common counter-stories in our dataset are first-person accounts from members of the targeted group, often addressed to the original poster and sometimes reusing the stereotype's language in order to invert or reject it. For example, a comment such as \textit{``They called us `lazy immigrants,' but my mother cleaned offices [...] so we could stay in school.''} repeats the stereotype while turning it into a testimony of labor and sacrifice. Keyword- or toxicity-based systems that are neither relational nor identity-sensitive may flag such comments as harmful even when they function as resistance \citep{gligoric-etal-2024-nlp, Gunturi2024LinguisticallyDA, Lee2024PeopleWS, field-etal-2021-survey}. We find some confirmation of this concern through a toxicity audit of our dataset. Using Google's Perspective API with a 0.5 toxicity threshold, 9.4\% of counter-stories were flagged as toxic, despite their role in responding to stereotypes, whereas 89.9\% of stereotype-containing posts fell below the threshold. This problem arises because counter-stories derive their meaning from context. They are \textit{relational}: a comment that appears to repeat a stereotype in isolation may be rejecting it in response to a parent post. They are also \textit{positional}: the same narrative can carry different social meaning depending on the speaker's relationship to the targeted group \citep{zhou2023cobraframes}. Moderation pipelines that are neither relational nor identity-sensitive will consistently misclassify counter-stories as harmful, thereby suppressing important forms of community responses that challenge stereotypes \citep{Fraser2016RethinkingTP, Lee2024PeopleWS, 10.1145/3351095.3372826}.\looseness=-1

Beyond moderation, our framework may also help surface authentic counter-stories so they are more likely to reach relevant audiences. Prior work has shown that a major barrier for counterspeakers is the perception that their efforts go unseen or have little impact \citep{Mun2024CounterspeakersPU, 10.1145/3745769}. Increasing the visibility of counter-stories may therefore help sustain counterspeakers' engagement while promoting solidarity and cohesion within targeted communities.\looseness=-1

\section{Conclusion}
\label{sec:conclusion}
Counter-storytelling is a critical tool for resisting racial stereotypes, yet its online practice remains poorly understood. We address this gap with a three-dimensional taxonomy grounded in narratology and critical race theory, capturing narrative position, audience, and identity. Using a multi-stage pipeline and LLM-assisted annotation, we construct a dataset of 25,549 Reddit posts across 615 communities, yielding 1,312 counter-story comments and 3,600 stereotypical posts. Our analysis reveals that counter-storytelling is diverse and relational, and we identify seven rhetorical roles that counter-storytellers commonly adopt. 

\section{Limitations}
\label{sec:limits}

Our study has several limitations. First, our analysis focuses on English-language Reddit communities discussing racial stereotypes.  Because our data is limited to English, the narrative forms, rhetorical conventions, and identity disclosures we observe may not generalize to counter-storytelling in other languages and cultural contexts, where norms around self-disclosure can differ. Our findings are also shaped by Reddit's threaded, long-form text format. Platform affordances may influence all three dimensions of our taxonomy: short-form video platforms such as TikTok may favor first-person testimony; quote-tweets and reply mechanisms on Twitter/X may shape whom a counter-story addresses; and private or semi-private forums may enable forms of in-group-directed storytelling that are less visible in public data. Counter-storytelling may likewise take different forms on other platforms with different interactional norms (e.g., Twitter, TikTok, private forums), in other linguistic and cultural contexts, or when responding to stereotypes about other marginalized identities such as gender, disability, religion, or sexuality.  Extending our framework across languages, platforms, and identity axes is therefore an important direction for future work.

Second, our dataset construction reflects methodological trade-offs. Our stereotype discovery pipeline combines lexicon-based retrieval, LLM-assisted labeling, and concept--carrier sentence expansion to identify posts containing stereotypes at scale. While this improves coverage beyond simple keyword matching, it may still miss subtler or community-specific stereotypes that do not align with our seed concepts, while LLM labeling may introduce false positives or biases in what is recognized as stereotypical content.

Third, identity annotations rely only on explicit self-disclosure in text. Our analysis therefore captures how users position their identities within the conversation, rather than the full demographic composition of participants.

\section{Ethical Considerations}
\label{sec:ethics}

The data we examine in this paper is public but sensitive. This tension leads to a series of tradeoffs between scientific transparency and reproducibility, privacy and agency of the Reddit users, and safety of our annotators and readers. Many of the stories and counter-stories in this study contain very personal information about the user including aspects of their identity. We determined that review for this study was not required by our institution's Internal Review Board (IRB), as users do not have an expectation of privacy, but the nature of this topic and data requires careful consideration.

\paragraph{Data decisions.} 

Both stereotype-containing posts and storytelling comments may contain distinctive personal details that could enable re-identification. To ensure counter-storytellers' privacy and agency,
we therefore do not release raw text, user identifiers, or post-comment pairs. However, to support research transparency while minimizing redistribution of sensitive narratives, we will make the annotated dataset available upon request for research purposes only. All requests will be reviewed with attention to the goals of the research, the intended use of the data, and safeguards for handling sensitive personal narratives. We publicly release the annotation codebook, pipeline code, race lexicon, and carrier sentences needed to support reproducibility.

All examples used in the paper were de-identified; no usernames or other identifying information are included.
Importantly, we never try to \textit{predict} identity features of the users and only use explicit disclosures (``\textit{As an Asian woman, ...}'') to categorize the texts; these public disclosures were intentional choices of the users and are significant to building our understanding of counter-storytelling.
We also minimize our data collection, and we do not collect additional data about these users; e.g., we do not track individual users across conversations and communities, and we do not attempt to construct profiles of individual users.

\paragraph{Risks.} We note two potential misuses: (1) the codebook could be mis-used to craft (manually or automatically) deceptive narratives, and (2) automated detection of counter-storytelling could be misused by bad actors to locate, monitor, or suppress resistance by marginalized communities.

\paragraph{Safety.} Our annotators (including the authors) and to a lesser extent, our readers, are exposed to extremely explicit  and toxic racist stereotypes through the data. Annotators discussed taking breaks and prioritizing their mental health. Unfortunately, the nature of the research questions posed in this work necessitate reading and contending with many different kinds of racial stereotypes, which often co-occur with misogynist and other upsetting language. We provide a warning at the beginning of the paper, and we minimize the toxicity of the examples shown in the paper, but given our research focus, we cannot avoid directly engaging with this content. We also acknowledge our positionality in this work: some authors belong to the stereotyped communities represented in our study, which informed our engagement with and interpretation of these narratives, while some authors do not belong to these groups.

\paragraph{Use of AI Assistance}
We used AI assistants for limited support during coding, analysis, and writing. All data decisions, annotations, citations, statistical analyses, interpretations, and final writing decisions were reviewed and made by the authors.

\section*{Acknowledgments}

We are especially grateful to Dr. Long Le Khac for the many generous discussions that helped shape and develop our annotation framework. We thank Teagan Johnson,  Advait Deshmukh, and colleagues at Carnegie Mellon University and the University of Colorado Boulder for helpful feedback throughout the project. We also thank the anonymous reviewers for their constructive comments.

\bibliography{custom}

\appendix

\section{Race Lexicon}
\label{sec:race_lexicon_appendix}

Our race lexicon comprises identity keywords (neutral terms for self-identification (e.g., ``Black,'' ``Asian American,'' ``Hispanic''), slurs (sourced from \url{https://slurs.info/slurs/}), and dog-whistles (sourced from \url{https://dogwhistles.allen.ai/}) for three racial groups. Tables~\ref{tab:lexicon_african}--\ref{tab:lexicon_asian} present example terms from each racial group; the complete lexicon is available in our supplementary materials.

\begin{table}[h]
\centering
\small
\begin{tabular}{p{2.1cm}p{4.0cm}}
\toprule
\textbf{Category} & \textbf{Example Terms} \\
\midrule
\textbf{Identity keywords} & 
black, african american, bipoc, african, african descent, afro-american \\
\midrule
\textbf{Dog-whistles} & 
jogger(s), ghetto(s), urban, welfare queen(s), thug(s), black-on-black crime, absent fathers, 13\%, 13/50 \\
\midrule
\textbf{Slurs} & 
[We include 600+ slurs in our lexicon but omit them here due to their highly offensive nature. Full list available upon request for research purposes.] \\
\bottomrule
\end{tabular}
\caption{Black/African race lexicon. Identity keywords: $n=9$; Dog-whistles: $n=28$; Slurs: $n=616$.}
\label{tab:lexicon_african}
\end{table}

\begin{table}[h]
\centering
\small
\begin{tabular}{p{2.1cm}p{4.0cm}}
\toprule
\textbf{Category} & \textbf{Example Terms} \\
\midrule
\textbf{Identity keywords} &
latino, latina, latinx, hispanic, mexican american, chicano, puerto rican, cuban, central american, south american \\
\midrule
\textbf{Dog-whistles} & 
secure the border, illegal aliens, criminal aliens, anchor baby/babies, stealing jobs, cartel, deport them all, press 2 for spanish, mexifornia \\
\midrule
\textbf{Slurs} & 
[We include 170+ slurs in our lexicon but omit them here due to their highly offensive nature. Full list available upon request for research purposes.] \\
\bottomrule
\end{tabular}
\caption{Hispanic race lexicon. Identity keywords: $n=15$; Dog-whistles: $n=24$; Slurs: $n=175$.}
\label{tab:lexicon_hispanic}
\end{table}

\begin{table}[h]
\centering
\small
\begin{tabular}{p{2.1cm}p{4.0cm}}
\toprule
\textbf{Category} & \textbf{Example Terms} \\
\midrule
\textbf{Identity keywords} & 
asian, asian American, east asian, south asian, southeast asian, indian, chinese, japanese, korean, vietnamese, filipino, pakistani, bangladeshi, indonesian \\
\midrule
\textbf{Dog-whistles} & 
model minority, china virus, kung flu, wuhan virus, foreigner, tiger mom, abolish birthright citizenship, ccp influence, academic cheats, tech job thieves, high iq, stem overlords \\
\midrule
\textbf{Slurs} & 
[We include 340+ slurs in our lexicon but omit them here due to their highly offensive nature. Full list available upon request for research purposes.] \\
\bottomrule
\end{tabular}
\caption{Asian race lexicon. Identity keywords: $n=19$; Dog-whistles: $n=25$; Slurs: $n=345$.}
\label{tab:lexicon_asian}
\end{table}

\section{Counter-Story Annotation Codebook}
\label{sec:codebook}

\begin{figure}[t]
 \definecolor{narrviolet}{RGB}{120, 60, 187}
    \definecolor{audorange}{RGB}{43, 105, 26}
    \definecolor{identblue}{RGB}{37, 150, 190}
    \includegraphics[
        width=\columnwidth,
        clip
    ]{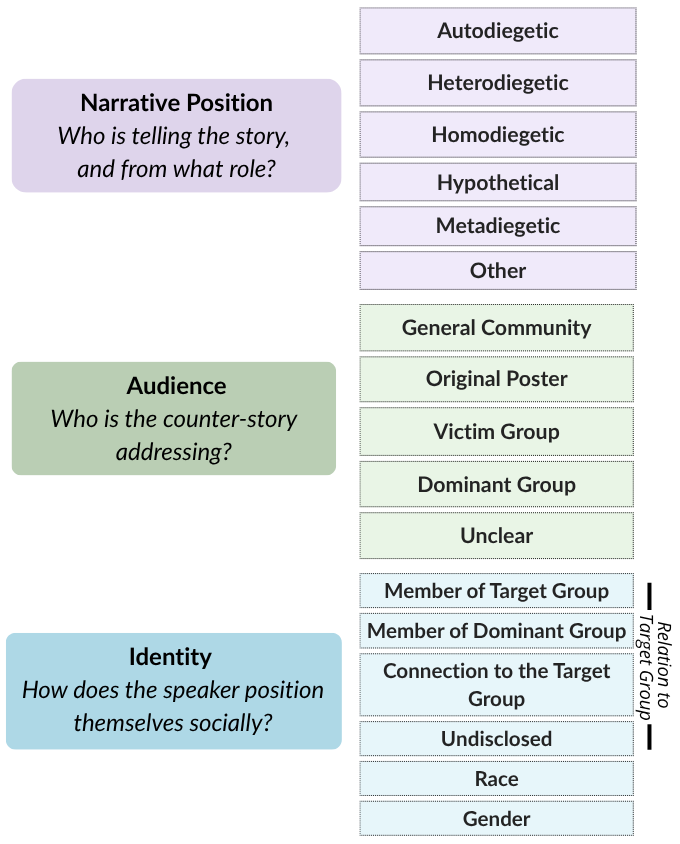}
    \caption{Our three-dimensional taxonomy for characterizing counter-stories:
    \textcolor{narrviolet}{\textbf{Narrative Position}} captures how the narrator formally relates to the story, \textcolor{audorange}{\textbf{Audience}} captures who the counter-story addresses; comments may address multiple audiences, and
    \textcolor{identblue}{\textbf{Identity}} captures how the speaker positions themselves relative to the stereotyped group, including self-disclosed demographic attributes.}
    \label{fig:taxonomy}
\end{figure}

Fig.~\ref{fig:codebook} displays the codebook used to annotate counter-narratives along three dimensions of our taxonomy (Fig.~\ref{fig:taxonomy}): narrative position, audience, and author identity.

\begin{figure*}[p]
\setlength{\fboxsep}{6pt}
\fbox{%
\begin{minipage}{\dimexpr\linewidth-2\fboxsep-2\fboxrule\relax}
    \fontsize{8}{9}\selectfont

    \textbf{1. NARRATIVE POSITION}

    \vspace{4pt}
    Captures the narrator's relationship to the story being told---whether the narrator stands outside the story world, appears within it, or recounts their own experience as its protagonist \citep{Genette1980NarrativeD, chatman1978story}.

    \vspace{4pt}
\begin{enumerate}[leftmargin=*, itemsep=3pt, topsep=4pt, label=\arabic*.]
        \item \textbf{Autodiegetic.}\newline
        \textit{Definition:} The narrator is simultaneously narrator and protagonist, recounting firsthand experiences of discrimination or resistance in the first person, most often as a member of the stereotyped group.\newline
        \textit{Example:} \textit{``As a Black diner, the waiter checked on the white couple beside us five times and never once on us.''}

        \item \textbf{Homodiegetic.}\newline
        \textit{Definition:} The narrator is present in the story as a witness rather than its protagonist. It could be:
        \begin{itemize}[noitemsep, leftmargin=*, topsep=2pt]
            \item a victim-group member witnessing discrimination toward others in their group,
            \item a dominant-group member recounting positive experiences with the victim group,
            \item a third-party observer of interactions between groups.
        \end{itemize}
        \textit{Example:} \textit{``I waited tables for years and watched servers breeze past the Black families in their section.''}

        \item \textbf{Heterodiegetic.}\newline
        \textit{Definition:} The narrator recounts stories they were not present for, spanning three subtypes:
        \begin{itemize}[noitemsep, leftmargin=*, topsep=2pt]
            \item \textit{Public exemplars}---public figures, historical events, or well-known cases.
            \item \textit{Second-hand}---stories heard through trusted sources.
            \item \textit{Fictional}---books, films, or other media invoked to challenge the stereotype.
        \end{itemize}
        \textit{Examples:}
        \begin{itemize}[noitemsep, leftmargin=*, topsep=2pt]
            \item (\textit{exemplar}) \textit{``Condoleezza Rice writes about a principal thanking her `for showing us that not all black people are stupid'.''}
            \item (\textit{second-hand}) \textit{``My coworker told me her landlord refused to rent to her the second he heard her accent over the phone.''}
            \item (\textit{fictional}) \textit{``Read Native Speaker; it captures growing up Asian American better than any argument.''}
        \end{itemize}

        \item \textbf{Metadiegetic.}\newline
        \textit{Definition:} The narrator tells a story within a story, embedding another person's (e.g., quoted or reported) account.\newline
        \textit{Example:} \textit{``My grandmother used to tell us about a waiter who said `we don't serve your kind here'...she'd describe how the whole room just went quiet.''}

        \item \textbf{Hypothetical.}\newline
        \textit{Definition:} The narrator constructs a counterfactual scenario, inviting the audience to reason through an alternative framing.\newline
        \textit{Example:} \textit{``Imagine a white teenager getting stopped and frisked walking home...would anyone call that `routine'?''}
    \end{enumerate}

    \vspace{3pt}
    \textbf{2. AUDIENCE}

    \vspace{3pt}
    Captures the implied primary recipient of the counter-story, since the same story can function differently depending on whom it addresses \citep{Fraser2016RethinkingTP, warner2000counterpublics}. A single comment can address multiple audiences.

    \begin{itemize}[noitemsep, leftmargin=*]
        \item \textbf{Original Poster (Direct Rebuttal).} Directly responds to the OP's stereotyping statement through personal confrontation.
        \item \textbf{Dominant Group.} Challenges or educates those who perpetuate the stereotype.
        \item \textbf{Victim Group (In-group).} Offers solidarity, guidance, or critique to in-group members.
        \item \textbf{General/Mixed Audience.} Speaks to the broader community; educational or contextual.
        \item \textbf{Ambiguous/Multiple.} The primary audience is unclear or addressed simultaneously.
    \end{itemize}

    \vspace{3pt}
    \textbf{3. AUTHOR IDENTITY}

    \vspace{4pt}
    Captures the narrator's social perspective. Identity comprises explicit self-disclosures, separating narrators who speak from lived membership in the stereotyped group from those who speak as witnesses or allies.

    \begin{itemize}[noitemsep, leftmargin=*]
        \item \textbf{Race:} White, Asian, Black, Hispanic, Other, Unknown.
        \item \textbf{Gender:} Man, Woman, Non-Binary, Trans, Unknown.
        \item \textbf{Member of Target Group:} Yes / No / Unknown.
        \item \textbf{Member of Dominant Group:} Yes / No / Unknown.
        \item \textbf{Personal Connection to Target Group:} Yes / No / Unknown (only if not a member of the target group).
    \end{itemize}

    \vspace{2pt}
    \textbf{COMMENTS THAT ARE \emph{NOT} COUNTER-NARRATIVES}

    \vspace{3pt}
    A counter-narrative must actively challenge the specific stereotype in the post. A comment is \emph{not} a counter-narrative if it:
    \begin{itemize}[noitemsep, leftmargin=*]
        \item agrees with, amplifies, or reinforces the stereotype (even with surface-level disagreement);
        \item is a pseudo-counter-narrative that appears to challenge the stereotype but ultimately reinforces it;
        \item changes the subject entirely or makes unrelated observations;
        \item offers general commentary about racism without specifically countering the post's stereotype; or
        \item contains no narrative (pure argument, data citation, or abstract meta-commentary).
    \end{itemize}

\end{minipage}%
}
\caption{Codebook for counter-story annotation.}
\label{fig:codebook}
\end{figure*}

\section{LLM Annotation Prompts}
\label{sec:appendix_prompts}

\subsection{Stereotype Detection}
\label{app:stereotype_prompt}
We show the few-shot prompt below; the zero-shot variant is identical with the
\textbf{Examples} block removed.

\begin{tcolorbox}[colback=gray!5!white, colframe=gray!75!black, title=System, breakable]
\small
You are an expert content analyst for racial stereotypes in online discourse. You are analyzing Reddit posts pre-filtered for racial/ethnic content. Your job is to determine whether the post expresses a racial stereotype.

\medskip\noindent\textbf{Definition.}
A racial stereotype is a claim attributed to a racial/ethnic group that is either:
(a) a generalization applied broadly to the group, OR \\
(b) an individual anecdote framed as representative of the group, OR \\
(c) humor, jokes, or slurs that dehumanize or demean a racial group. \\
Stereotypes can be negative, positive (``model minority''), or use coded language (``those neighborhoods'').

\medskip\noindent\textbf{Key Rule.}
The author must ENDORSE or ASSERT the stereotype. Reporting, quoting, or refuting someone else's stereotype does NOT count --- even if the post is about stereotypes.

\medskip\noindent\textbf{Classify YES when:}
\begin{itemize}[label={--}, leftmargin=*, nosep]
  \item The author generalizes about a racial/ethnic group, explicitly or implicitly
  \item Individual/family anecdotes are framed as ``typical'' or representative of the group
  \item Positive-framed stereotypes (model minority, ``naturally athletic'')
  \item Coded/dog-whistle language when the racial group is clear from context
  \item In-group authors count --- judge the content, not the author's identity
  \item Hedged claims count if the underlying assertion is endorsed
  \item Jokes whose punchline relies on racial slurs or dehumanizes a racial group count even without explicit trait attribution
\end{itemize}

\medskip\noindent\textbf{Classify NO when:}
\begin{itemize}[label={--}, leftmargin=*, nosep]
  \item Author is reporting, quoting, analyzing, or refuting a stereotype
  \item Post cites statistics/data without claiming inherent group traits
  \item Post describes a specific individual without framing them as representative
  \item Fictional/hypothetical contexts without stereotypical claims
  \item Sarcastic reversals clearly mocking the stereotype
\end{itemize}

\medskip\noindent\textbf{Examples.}
\begin{itemize}[label={}, leftmargin=*, nosep]
  \item \textbf{Example 1 --- YES:} ``Why are Black people always so loud at restaurants lol'' \\
  \texttt{\{"stereotype":"Yes", "target\_group":"Black", "stereotype\_type":"cultural", "rationale":"Explicit generalization.", "certainty":9\}}
  \item \textbf{Example 2 --- NO:} ``My professor talked about how Asian students are stereotyped as model minorities, it was an eye-opening lecture about implicit bias.'' \\
  \texttt{\{"stereotype":"No", "target\_group":"None", "stereotype\_type":"none", "rationale":"Reports critically; author does not endorse.", "certainty":9\}}
  \item \textbf{Example 3 --- YES:} ``Every Black guy I've met plays basketball, it's just a fact lol'' \\
  \texttt{\{"stereotype":"Yes", "target\_group":"Black", "stereotype\_type":"cultural", "rationale":"Anecdote asserted as general fact.", "certainty":9\}}
  \item \textbf{Example 4 --- NO:} ``My dad keeps saying all Mexicans are criminals but he's never met one. Tired of this ignorance.'' \\
  \texttt{\{"stereotype":"No", "target\_group":"None", "stereotype\_type":"none", "rationale":"Quotes and refutes.", "certainty":9\}}
  \item \textbf{Example 5 --- YES:} ``I don't want to be racist but there's a reason property values drop when Black people move in...'' \\
  \texttt{\{"stereotype":"Yes", "target\_group":"Black", "stereotype\_type":"economic", "rationale":"Coded reference; hedging confirms awareness.", "certainty":8\}}
  \item \textbf{Example 6 --- NO:} ``Oh yeah, all Asians are good at math and never have feelings outside of studying, right?'' \\
  \texttt{\{"stereotype":"No", "target\_group":"None", "stereotype\_type":"none", "rationale":"Sarcastic reversal mocking the stereotype.", "certainty":8\}}
\end{itemize}

\medskip\noindent\textbf{Certainty Scale.}
\begin{itemize}[label={$\cdot$}, leftmargin=*, nosep]
  \item 9--10: Unambiguous
  \item 7--8: Clear but requires interpretation
  \item 5--6: Ambiguous about endorsement or scope
  \item 1--4: Not a stereotype
\end{itemize}

\medskip\noindent\textbf{Analysis Steps.}
\begin{enumerate}[nosep]
  \item Target group
  \item Trait/claim attributed --- if none, check for slurs or dehumanization
  \item Endorsement: asserting vs.\ reporting?
  \item Verdict
\end{enumerate}

\medskip\noindent Output ONLY valid JSON: \\
{\footnotesize\ttfamily\raggedright
\{"reasoning":"<2-3 sentences>",\\
"stereotype":"Yes/No",\\
"target\_group":\\
"<Black/Asian/Mexican/White/Other/None>",\\
"stereotype\_type":\\
"<dehumanization/criminality/intelligence/\\
appearance/cultural/economic/other/none>",\\
"rationale":"<1-2 sentences>",\\
"certainty":<1-10>\}
\par}
\normalsize
\end{tcolorbox}

\begin{tcolorbox}[colback=gray!5!white, colframe=gray!75!black, title=User, breakable]
\small
\textbf{Subreddit:} \texttt{r/\{subreddit\}} \\
\textbf{Post Title:} ``\{title\}'' \\
\textbf{Original Post:} ``\{post\_text\}''
\normalsize
\end{tcolorbox}

\subsection{Counter-Story Annotation}
\label{sec:appendix_prompts_counter_narratives}
For counter-story annotation, we experimented with both zero-shot and few-shot prompting and
observed no significant difference in agreement with the human labels; we therefore use the
codebook-guided prompt below.

\begin{tcolorbox}[colback=gray!5!white, colframe=gray!75!black, title=System, breakable]
\small
You are an expert annotator for counter-narrative research on racial stereotypes in online discourse.

\medskip\noindent\textbf{What is a Counter-Narrative?}
A counter-narrative is a story or response that:
\begin{itemize}[label={--}, leftmargin=*, nosep]
  \item Opposes or subverts a dominant/majoritarian narrative (here: a racial stereotype)
  \item Centers the lived experience or dignity of the targeted group
  \item Reframes the representation in a way that undermines the stereotype's legitimacy
\end{itemize}
Counter-narratives oppose hate/stereotype content through strategies like presenting contradicting facts or lived experience, pointing out hypocrisy or logical contradiction, humanizing the targeted group, or reframing the narrative from the targeted group's perspective.

\medskip\noindent\textbf{Key rule.}
A counter-narrative must actively challenge the specific stereotype in the post, not merely agree with or amplify the stereotype, change the subject entirely, make unrelated observations, or add general commentary about racism without specifically countering the stereotype. Comments that agree with or reinforce the stereotype, or pseudo-counter-narratives that appear to challenge but ultimately reinforce the stereotype, are NOT counter-narratives.

\medskip\noindent\textbf{Annotation Dimensions.}
\begin{itemize}[label={}, leftmargin=*, nosep]
\item \textbf{0a. CONTAINS NARRATIVE?}\\
- Does the comment include a story, personal experience, observation, or retelling of events? \\
- Note that pure arguments, data citations, or abstract meta-commentary WITHOUT story elements = No. \\
- Note that a comment can contain multiple narratives on different topics---that still counts as Yes. \\
\textit{Yes / No}
\item \textbf{0b. IS COUNTER-NARRATIVE?} Does this comment challenge, contradict, humanize, or provide an alternative perspective to the racial stereotype in the post? Mark No if it agrees with/reinforces the stereotype, does not meaningfully challenge and counter it, doesn't contain a narrative, or makes general anti-racism commentary without addressing the post's stereotype. \textit{Yes / No}
\item \textbf{1. NARRATIVE POSITION} (if 0a = No, use ``N/A''):
\begin{itemize}[label={$\cdot$}, leftmargin=1.5em, nosep]
  \item \textit{Autodiegetic}---narrator is protagonist with direct personal experience
  \item \textit{Homodiegetic}---narrator present as witness/observer, not protagonist
  \item \textit{Heterodiegetic}---third-person stories narrator was NOT present for
  \item \textit{Metadiegetic}---story within a story, nested/multiple narrative levels
  \item \textit{Hypothetical}---hypothetical scenarios or thought experiments
  \item \textit{Other}---narrative position present but doesn't fit above
  \item \textit{N/A}---no narrative position
\end{itemize}
\item \textbf{2. NARRATIVE SUBTYPE} (only if Heterodiegetic, else ``N/A''; comma-separated if multiple):
\begin{itemize}[label={$\cdot$}, leftmargin=1.5em, nosep]
  \item \textit{Second-hand}---stories from people narrator knows
  \item \textit{Exemplar/Celebrity}---stores about historical figures, celebrities, public cases
  \item \textit{Fictional}---stories related to fiction, media, books, movies
\end{itemize}
\item \textbf{3. AUDIENCE} (all that apply, comma-separated):
\begin{itemize}[label={$\cdot$}, leftmargin=1.5em, nosep]
  \item \textit{Dominant Group} (challenging/educating those who perpetuate stereotypes)
  \item \textit{Victim Group} (solidarity, guidance, or critique to in-group)
  \item \textit{Original Poster} (directly responding to OP)
  \item \textit{General/Mixed} (broader Reddit community)
  \item \textit{Unclear}
\end{itemize}
\item \textbf{4. IDENTITY -- RACE} (stated or clearly implied): White / Black / Asian / Hispanic / Other / Unknown
\item \textbf{5. IDENTITY -- GENDER} (stated or clearly implied): Man / Woman / Non-Binary / Trans / Unknown
\item \textbf{6. MEMBER OF TARGET GROUP?} Is the narrator a member of the stereotyped group? Yes / No / Unknown
\item \textbf{7. MEMBER OF DOMINANT GROUP?} Is the narrator a member of the dominant (typically white/majority) group? Yes / No / Unknown
\item \textbf{8. CONNECTION TO TARGET GROUP} (only if NOT a member of target group): \\
Is the narrator a personal connection (family, partner, close friend) to the target group? Yes / No / Unknown / N/A
\item \textbf{9. RATIONALE} (step-by-step reasoning---do this FIRST, before fixing your verdicts): \\
(a) What is the specific racial stereotype in the post? \\
(b) What is the commenter actually saying/doing? Summarize the comment's stance. \\
(c) Is there a narrative (personal experience, retelling, observation)? What kind? \\
(d) Does the comment meaningfully challenge the stereotype in (a)? How, or why not? \\
(e) Any edge cases, ambiguity, sarcasm, or pseudo-counter-narrative features? \\
Be concrete, reference actual content from the comment.
\item \textbf{10. CERTAINTY} (1--10), for the \texttt{is\_counter\_narrative} verdict: 1--3 = very unsure; 4--6 = moderate; 7--9 = high; 10 = unambiguous
\end{itemize}

\medskip\noindent Output ONLY a valid JSON object. Fill \texttt{rationale} FIRST (chain-of-thought), then use that reasoning to fill the verdicts:\\
\texttt{\{"rationale":"...", \\
"contains\_narrative":"Yes/No", \\
"is\_counter\_narrative":"Yes/No", \\
"narrative\_position":"...", \\
"narrative\_subtype":"...", \\
"audience":"...", \\
"identity\_race":"...", \\
"identity\_gender":"...", \\
"member\_target\_group":"...", \\
"member\_dominant\_group":"...", \\
"connection\_target\_group":"...", \\
"certainty":1-10\}}
\normalsize
\end{tcolorbox}

\begin{tcolorbox}[colback=gray!5!white, colframe=gray!75!black, title=User, breakable]
\small
\textbf{Subreddit:} \texttt{r/\{subreddit\}} \\
\textbf{Post Title:} ``\{title\}'' \\
\textbf{Original Post} (contains racial stereotype about \texttt{\{target\_group\}}): ``\{post\_text\}'' \\
\textbf{Comment to analyze:} ``\{comment\_text\}''
\normalsize
\end{tcolorbox}

\begin{figure*}[t]
    \centering
    \includegraphics[width=0.8\textwidth]{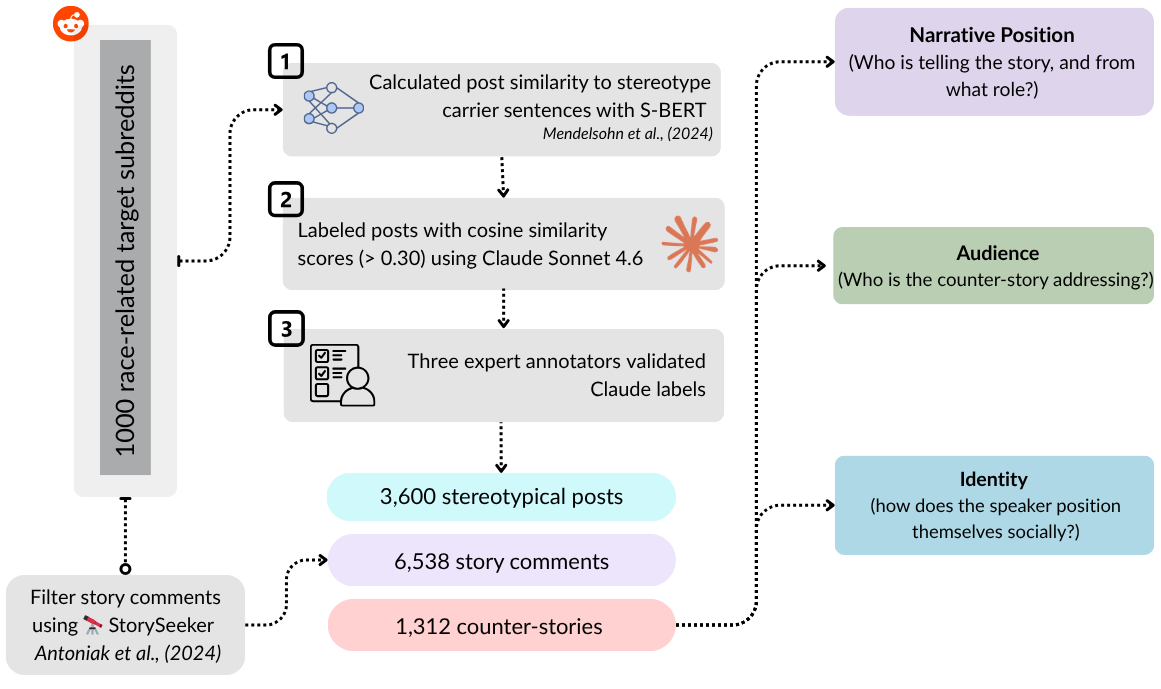}
    \caption{Overview of the multi-stage data filtering pipeline.}
    \label{fig:pipeline}
\end{figure*}

\section{Stereotype Concepts and Carrier Sentences}
\label{sec:concepts_stereotpe_carrier_sentences}
We follow the methodological approach outlined in \cite{mendelsohn2025peoplefloods} for generating carrier sentences associated with stereotype \textit{concepts}. Similar to their framework, we compile source stereotype sentences for each concept from literature (see Table~\ref{tab:literature} for more details) per race group and use them as the foundation for generating new implicit and explicit carrier sentences. To generate these sentences, we provided GPT--4o with the standardized prompt shown below.

\begin{tcolorbox}[colback=gray!5!white, colframe=gray!75!black, title=System, breakable]
\small
A \textbf{concept} is the stereotype category or stereotypical association to be expressed. A \textbf{carrier sentence} is a short, generic sentence that expresses that concept with minimal additional semantic detail. Given a concept and the example stereotypes, generate 5 implicit and 5 explicit versions of the stereotypical carrier sentences per concept using the example sentences provided. Limit it to 5 and not more than that. Do not generate any other information or reasoning.
\normalsize
\end{tcolorbox}
\begin{tcolorbox}[colback=gray!5!white, colframe=gray!75!black, title=User, breakable]
\small
\textbf{Concept:} \texttt{\{concept\}} \\
\textbf{Example sentences:} \texttt{\{source\_sentences\}}
\normalsize
\end{tcolorbox}

\begin{table}[t]
\centering
\footnotesize
\setlength{\tabcolsep}{4pt}
\renewcommand{\arraystretch}{1.3}
\begin{tabularx}{\columnwidth}{lX}
\toprule
\textbf{Group} & \textbf{Sources} \\
\midrule
Black    & \cite{BannerHaley2011TheCO}; \cite{HarrisPerry2011SisterCS}; \cite{Tyson2019ReviewOR}; \cite{Taylor2012DarkestAB}; \cite{Tucker2007LockstepAD}; \cite{Bogle1973TomsCM} \\
\addlinespace
Asian    & \cite{Li2015EllenDW}; \cite{Roediger1999OrientalsAA}; \cite{Markus2010DoingR}; \cite{Majumdar2016InSS} \\
\addlinespace
Hispanic & \cite{NezMchiri2012TheLT}; \cite{Bebout2016WhatDT}; \cite{Alonzo2009BadmenBA}; \cite{Limn1973StereotypingAC} \\
\bottomrule
\end{tabularx}
\caption{Sources from which we drew dominant-stereotype concepts. Concepts and carrier sentences were constructed from stereotype expressions documented in these works.}
\label{tab:literature}
\end{table}

\begin{table}[t]
\centering
\footnotesize
\setlength{\tabcolsep}{4pt}
\renewcommand{\arraystretch}{1.3}
\begin{tabularx}{\columnwidth}{@{}lX@{}}
\toprule
\textbf{Group} & \textbf{Concepts} \\
\midrule
Black (20) & personality, crime, music, athletic, skills, appearance, socioeconomic, family relations, intelligence, welfare/laziness, absent fathers, AAVE mockery, angry, anti-intellectual, colorism, food (chicken/watermelon), gang violence, kinky hair, hypersexualization, racial assimilation \\
\addlinespace
Asian (27) & model minority, academics, tiger mom, language barrier, perpetual foreigner, appearance, dehumanization, COVID blame, eating dogs/cats, emasculation (men), fetishization (women), ladyboy trope, martial arts, orientalist, asian drivers, socioeconomic, work ethic, family structure, SE-Asian maid/nanny, SE-Asian mail-order bride, S.Asian IT scammer, S.Asian accent, S.Asian arranged marriage, S.Asian body hair, S.Asian caste, S.Asian curry smell, Vietnam-war tropes \\
\addlinespace
Hispanic (19) & immigration, crime, socioeconomic, economic impact (job-stealing), sexualization, culture, drug cartel, lazy/siesta, anchor babies, machismo, catholic large family, cuban hardliner, latino welfare, narco aesthetics, puerto rican violence, spanglish/uneducated, spicy latina, stealing jobs, wetback-coded \\
\bottomrule
\end{tabularx}
\caption{Full inventory of the 66 stereotype concepts (20 Black, 27 Asian, 19 Hispanic). Each concept has 10 carrier sentences (5 implicit, 5 explicit). The complete carrier-sentence set is released with our code.}
\label{tab:concept_inventory}
\end{table}

\begin{table*}[p]
\centering
\small
\begin{tabular}{lp{6.5cm}p{5cm}}
\toprule
\textbf{Concept} & \textbf{Implicit} & \textbf{Explicit} \\
\midrule
\multicolumn{3}{c}{\textbf{African}} \\
\midrule
Personality & They always seem to avoid work. & They are lazy. \\
 & You can never talk to them without a fight starting. & They are angry. \\
 & You'll hear them before you see them. & They are so loud. \\
 & Funny how they always pick the same political side. & They are racist. \\
 & They don't exactly fit into polite society. & They're uncivilized. \\
\midrule
Crime & They always seem to be around when something illegal happens. & They are involved in shooting activities. \\
 & You hear about them in crime reports all the time. & They will steal anything. \\
 & Their neighborhoods aren't exactly the safest. & They are more likely to commit crime. \\
 & They tend to get into trouble with the law more often. & They do a lot of drugs. \\
 & It's not surprising when they're caught with drugs. & They commit 50\% of violent crime. \\
\midrule
Music & You can always count on them to carry a tune. & Hip Hop/Rap is their culture. \\
 & Their parties are never quiet—there's always music. & They sing really well. \\
 & It's no surprise they nailed the rhythm section. & They are into music. \\
 & I figured they'd be great at freestyle. & They all have excellent rhythm. \\
 & Of course they knew every word to the rap song. & They have music running in their blood. \\
\midrule
Skills & They never sit out during sports. & They are naturally athletic. \\
 & You'll find them on every winning team. & They are good at sports. \\
 & They make everything look effortless on the field. & They excel in physical activities. \\
 & Coaches always want them first. & They dominate in every game. \\
 & It's like they have sports in their blood. & They're born to be athletes. \\
\midrule
Appearance & You can always tell them apart in a crowd. & They all look dark. \\
 & They don't quite blend in, do they? & They all have dark skin. \\
 & There's something intimidating about the way they look. & They all look big. \\
 & You know who they are just by their build. & They look tall and huge. \\
 & They have that ``wild'' look to them. & They are primitives or beasts. \\
\bottomrule

\end{tabular}
\caption{Implicit and explicit carrier sentences for five representative
Black stereotype concepts (of 20). The full set of 660 carrier
sentences across all 66 concepts is released with our code.}
\label{tab:carriers_black}
\end{table*}

\begin{table*}[p]
\centering
\small
\begin{tabular}{lp{6.5cm}p{5cm}}
\toprule
\textbf{Concept} & \textbf{Implicit} & \textbf{Explicit} \\
\midrule
\multicolumn{3}{c}{\textbf{Asian}} \\
\midrule
Model Minority & Of course she aced the math test. & They are really good at math. \\
 & He's the quiet type. & They are well behaved. \\
 & I bet their kid is doing calculus. & They are calm. \\
 & You never have to worry about them. & They are reserved. \\
 & He doesn't speak up much. & They are so quiet. \\
\midrule
Dehumanization & They look like they came from Chinatown. & They are just chinks. \\
 & I can't even tell them apart. & Chinks don't belong here. \\
 & They're always speaking in gibberish. & No one cares about chinks. \\
 & They're basically robots when they work. & Chinks aren't real Americans. \\
 & Just another face in the crowd. & Treat them like the chinks they are. \\
\midrule
Socioeconomic & You can count on them for low-cost labor. & They work for cheap. \\
 & The stuff they sell falls apart quickly. & They are cheap labor. \\
 & I wouldn't trust those sketchy websites. & They make cheap quality products. \\
 & Hard to tell if their products are real. & Their websites are fake. \\
 & Their businesses seem too good to be true. & They don't produce legitimate products. \\
\midrule
Orientalism & Their dinner smells like a dojo. & Their food is exotic. \\
 & I wouldn't mess with them. & They are Samurai and Ninja peoples. \\
 & Feels like an old martial arts movie. & They all know kung fu. \\
 & They learned from Samurai ancestors. & Their culture is full of martial arts. \\
 & Their recipes are strange yet magical. & They are obsessed with anime. \\
\midrule
Drivers & You can tell when one is driving. & Their women are bad drivers. \\
 & They have no concept of road rules. & They always drive rashly. \\
 & Driving with them feels like a gamble. & Their women can't handle cars. \\
 & Their women hesitate at every turn. & They drive too fast and dangerously. \\
 & Always a thrill ride for the wrong reasons. & Their women shouldn't drive. \\
\bottomrule
\end{tabular}
\caption{Implicit and explicit carrier sentences for five representative
Asian stereotype concepts (of 27). The full set of 660 carrier sentences
across all 66 concepts is released with our code.}
\label{tab:carriers_asian}
\end{table*}

\begin{table*}[p]
\centering
\small
\begin{tabular}{lp{6.5cm}p{5cm}}
\toprule
\textbf{Concept} & \textbf{Implicit} & \textbf{Explicit} \\
\midrule
\multicolumn{3}{c}{\textbf{Hispanic}} \\
\midrule
Immigration & They threaten our national security. & They are crossing the border illegally. \\
 & They show up without papers. & They don't know any English. \\
 & Their children are born here for citizenship. & They are a threat to national security. \\
 & They don't even try to learn the language. & They have anchor babies. \\
 & They're always finding ways around the system. &  \\
\midrule
Crime & You always hear sirens in their neighborhoods. & They run with gang members. \\
 & They hang out in sketchy parts of town. & They are into gang related activity. \\
 & Trouble just seems to follow them. & They are criminal, corrupt and lawless. \\
 & Their apartments are always full of people. & They are drug dealers. \\
 & You can tell something illegal is going on. & They lash out violently. \\
\midrule
Socioeconomic & You'll mostly find them cleaning. & They only do dirty jobs. \\
 & They tend to stick to manual labor. & They only get jobs at construction sites. \\
 & They usually don't have desk jobs. & They are housekeepers or maids. \\
 & They're keeping the place running. & They don't work in professional roles. \\
 & It's rare to see them in suits. & They can't afford higher education. \\
\midrule
Economic Impact & It's been harder to find stable work. & They're taking our jobs. \\
 & The paycheck doesn't stretch like it used to. & They're taking our manufacturing jobs. \\
 & The neighborhood's really changed. & They're taking our money. \\
 & People like us can't catch a break. & They're killing us. \\
 & Jobs disappeared right when they showed up. & They're ruining our livelihoods. \\
\midrule
Sexualization & They turn heads wherever they go. & They are known for being hypersexual. \\
 & There's something provocative about them. & They always wear revealing clothes. \\
 & They don't need makeup to stand out. & Everyone thinks they look exotic. \\
 & You can tell they like the attention. & They're always showing too much skin. \\
 & Their style speaks louder than words. & They're considered overly seductive. \\
\bottomrule
\end{tabular}
\caption{Implicit and explicit carrier sentences for five representative
Hispanic stereotype concepts (of 19). The full set of 660 carrier sentences
across all 66 concepts is released with our code.}
\label{tab:carriers_hispanic}
\end{table*}

\section{Additional Information about the Dataset}
\label{appendix:additional_results}

\subsection{False-Negative Audit for Missed Counter-Stories}
\label{appendix:fn_audit}
We conducted a small-scale false-negative audit focusing on two likely sources of missed counter-stories: (i) story-comments under stereotype posts that were labeled as non-counter-stories by the LLM, and (ii) story-comments under posts scoring just below the stereotype retrieval threshold (0.30). We manually inspected approximately 100 comments from these audit pools and found only $\sim$2\% missed counter-stories, providing some evidence that missed counter-stories are relatively uncommon. The missed cases tended to challenge stereotypes indirectly. For example, one countered an \textit{``Asian students cheat''} stereotype by recounting widespread cheating at the commenter's own school (rather than explicitly addressing it), reframing the behavior as situational rather than racial; Sonnet correctly classified this comment as a first-person narrative but failed to recognize that the story challenges a stereotype. Comprehensive false-negative estimation remains challenging because counter-stories are inherently sparse and require jointly identifying a stereotype, a story, and a countering response.

\subsection{Fiction and Media as Counter-Storytelling Resources}
\label{appendix:fiction_examples}

While most counter-stories in our dataset draw on lived experience, we also observed a small number of cases in which writers referenced fictional, literary, or media texts when responding to stereotypes. We do not analyze this phenomenon systematically, but include these examples to document an additional form of counter-storytelling present in the dataset: counter-storytelling may also operate through shared cultural texts, where fiction and media become resources for countering stereotypes.

\smallskip\noindent
\textbf{Example 1: Films.}
Responding to anti-BLM claims, one commenter invokes \textit{Nightcrawler} as a social analogy for unequal media attention given to certain racial groups: ``\textit{Renee Russo tells Jake Gyllenhaal that she doesn't want footage of a black or latino person dying, she wants news stories about affluent white people being killed}.'' The commenter uses this example to argue that ``\textit{Black Lives Matter}'' exists because Black people's suffering is often ignored or given less attention.

\smallskip\noindent
\textbf{Example 2: Parable.}
Responding to the claim that ``\textit{some peoples' (asians) cultural values are superior},'' a commenter retells the \textit{Cozumel fisherman} parable, in which a businessman urges a fisherman to expand his work only to eventually return to the simple life he already has. The parable is used to reject the premise of cultural hierarchy: ``\textit{You're ascribing your own cultural values and beliefs and saying that every other culture should share those. They're not necessarily better, they're just different}.''

\smallskip\noindent
\textbf{Example 3: Literary Fiction.}
In response to stereotypes around Asian masculinity, commenters recommend novels such as \textit{Native Speaker}, which offers ``\textit{great insights on what it means to grow up yellow in America},'' and \textit{The Raj Quartet}, where ``\textit{a white British author created a masculine Indian man as a romantic partner for a white English woman}.''

\subsection{Intersectional Stereotype Concepts}
\label{appendix:intersectional_stereotypes}
Our inventory also includes dedicated race$\times$gender (intersectional) concepts (e.g., Asian-male emasculation, Asian-female fetishization, Black hypersexualization, and Hispanic \emph{spicy latina} and \emph{machismo}). The carrier-sentence method detects such stereotypes directly: for example, the post ``\textit{Asian men get mocked in western countries in just about everything: height, frame, face\ldots even Asian women don't want us}'' was retrieved by the Asian-male emasculation concept at cosine $0.62$, well above our pipeline's $0.30$ threshold. A systematic analysis of how intersectional framing shapes counter-storytelling is left to future work.

\subsection{Hispanic-Targeted Content}
\label{appendix:hispanic_content}
\begin{figure*}[t]
\centering
\includegraphics[width=0.7\textwidth]{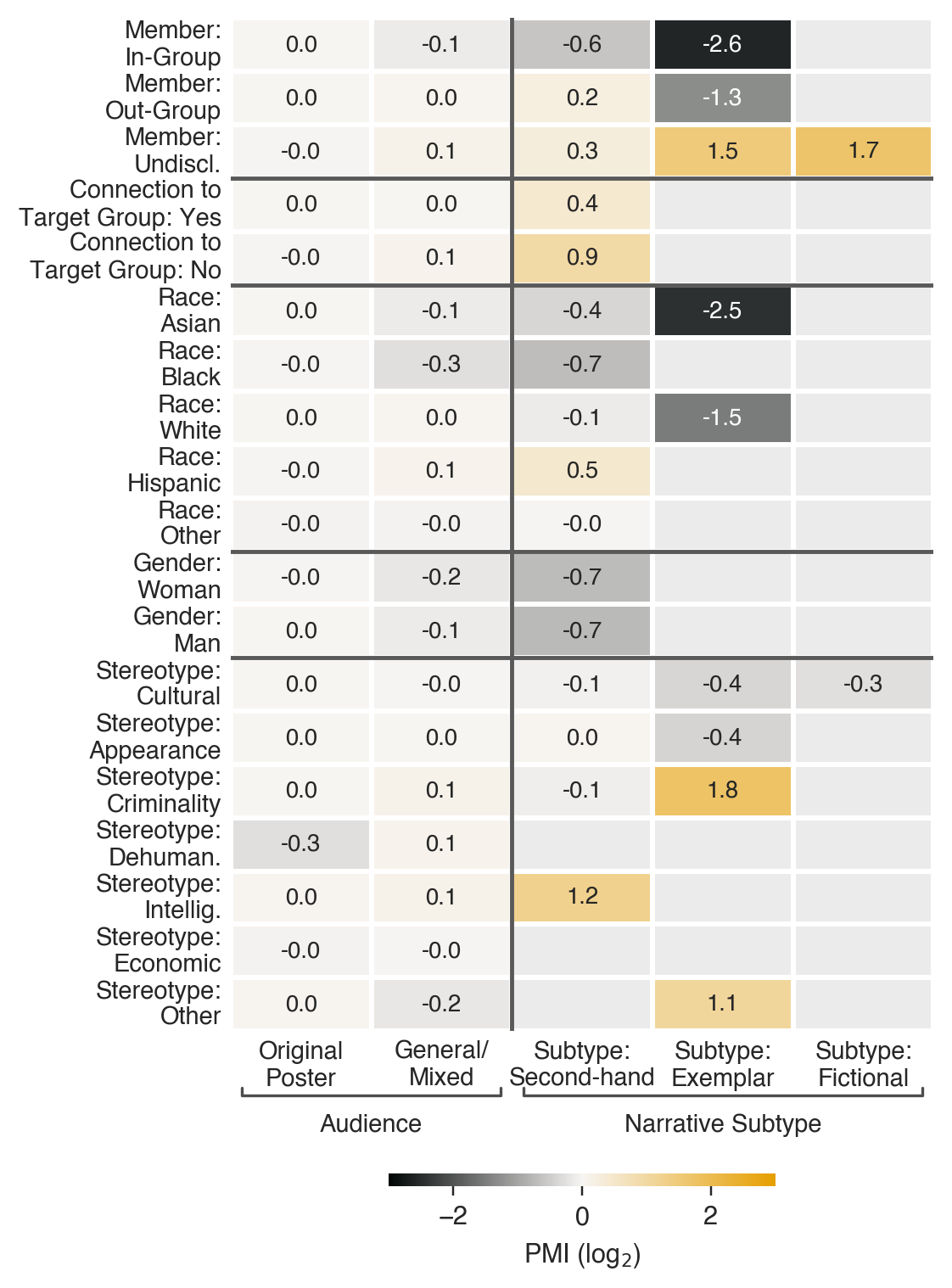}
\caption{Complete pointwise mutual information (PMI, $\log_2$) between counter-story
features and the audience and narrative-subtype categories not shown in the main
figures, run over the 1{,}312 counter-stories, extending Figures~\ref{fig:pmi_stereotype_full} and~\ref{fig:pmi_identity_full}. Rows cover the
writer's group membership, \emph{connection to the target group} (a personal tie without
membership), disclosed race and gender, and stereotype type; columns cover the remaining
audience categories (original poster, general/mixed) and the heterodiegetic narrative
subtypes (second-hand, exemplar, fictional). Orange cells indicate feature pairs that co-occur more than chance. Gray cells denote pairs with fewer than five co-occurrences ($n<5$), which we omit to avoid unstable estimates from sparse pairs.}
\label{fig:pmi_appendix}
\end{figure*}

Hispanic-targeted content is sparsely represented in our data (71 stereotype
posts and 34 counter-stories, $\sim$2\% of each category). These counts are too small for reliable PMI values or regression coefficients, so we exclude the Hispanic group from all cross-group comparisons in the main text and report its statistics here for completeness only.
Table~\ref{tab:stereotype_dist} reports the stereotype-type distribution for Hispanic-targeted posts, and the ``Race: Hispanic'' rows in Fig.~\ref{fig:pmi_appendix} and
Table~\ref{tab:race_position} give the corresponding narrative-position and subtype breakdowns.

\begin{table}[t]
\centering
\small
\setlength{\tabcolsep}{2.9pt}
\renewcommand{\arraystretch}{1.1}
\begin{tabular}{lrrrrr}
\toprule
\textbf{Stereotype type} & \textbf{Auto} & \textbf{Hete} & \textbf{Homo} & \textbf{Hypo} & \textbf{Meta} \\
\midrule
Cultural (n=1{,}018)   & 63.5 & 11.3 & 13.1 & 2.5  & 9.4  \\
Appearance (n=105)     & 61.9 & 7.6  & 14.3 & 1.9  & 14.3 \\
Criminality (n=83)     & 39.5          & 37.0 & 8.6  & 11.1 & 3.7  \\
Intelligence (n=33)    & 51.5          & 27.3 & 6.1  & 9.1  & 6.1  \\
Economic (n=25)        & 40.0          & 24.0 & 16.0 & 8.0  & 12.0 \\
Dehumanization (n=16)  & 43.8          & 31.2 & 12.5 & 12.5 & 0.0  \\
Other (n=32)           & 50.0          & 25.0 & 6.2  & 3.1  & 15.6 \\
\bottomrule
\end{tabular}
\caption{Stereotype type $\times$ narrative position among counter-stories (row \%; $n$ per type shown). Raw proportions underlying the associations in Fig.~\ref{fig:pmi_stereotype_full}.}
\label{tab:stereotype_position}
\end{table}

\definecolor{typeblue}{HTML}{3F6FA3}
\definecolor{targetteal}{HTML}{2E8B7F}

\begin{table*}[t]
\centering
\small
\setlength{\tabcolsep}{3pt}
\renewcommand{\arraystretch}{1.1}
\resizebox{\textwidth}{!}{%
\begin{tabular}{lrrrrrrrrrrrrrr}
\toprule
& & & & \multicolumn{7}{c}{\textit{Stereotype Type}} & \multicolumn{4}{c}{\textit{Target Group}} \\
\cmidrule(lr){5-11} \cmidrule(lr){12-15}
\textbf{Domain} & \textbf{\#Subs} & \textbf{\#Stereo.} & \textbf{Stereo.\%} & \textbf{Cult.} & \textbf{App.} & \textbf{Crim.} & \textbf{Deh.} & \textbf{Int.} & \textbf{Eco.} & \textbf{Oth.} & \textbf{Asian} & \textbf{Black} & \textbf{Hisp.} & \textbf{Other} \\
\midrule
\textit{Culture/Identity} & 139 & 988 & 18.5\% & \cellcolor{typeblue!55}85\% & \cellcolor{typeblue!20}4\% & \cellcolor{typeblue!10}4\% & \cellcolor{typeblue!20}2\% & \cellcolor{typeblue!20}2\% & \cellcolor{typeblue!10}1\% & \cellcolor{typeblue!10}2\% & \cellcolor{targetteal!55}40\% & \cellcolor{targetteal!10}10\% & \cellcolor{targetteal!10}1\% & \cellcolor{targetteal!55}50\% \\
\textit{Advice/Education/Work} & 91 & 796 & 12.0\% & \cellcolor{typeblue!55}73\% & \cellcolor{typeblue!35}10\% & \cellcolor{typeblue!20}6\% & \cellcolor{typeblue!10}2\% & \cellcolor{typeblue!35}3\% & \cellcolor{typeblue!20}2\% & \cellcolor{typeblue!20}4\% & \cellcolor{targetteal!55}41\% & \cellcolor{targetteal!20}19\% & \cellcolor{targetteal!20}2\% & \cellcolor{targetteal!35}39\% \\
\textit{Stories} & 28 & 420 & 16.7\% & \cellcolor{typeblue!55}70\% & \cellcolor{typeblue!35}10\% & \cellcolor{typeblue!35}6\% & \cellcolor{typeblue!35}8\% & \cellcolor{typeblue!10}1\% & \cellcolor{typeblue!20}1\% & \cellcolor{typeblue!20}3\% & \cellcolor{targetteal!35}30\% & \cellcolor{targetteal!35}31\% & \cellcolor{targetteal!35}2\% & \cellcolor{targetteal!20}37\% \\
\textit{Debate/Opinion} & 15 & 384 & 34.5\% & \cellcolor{typeblue!20}58\% & \cellcolor{typeblue!20}6\% & \cellcolor{typeblue!55}18\% & \cellcolor{typeblue!20}5\% & \cellcolor{typeblue!55}4\% & \cellcolor{typeblue!35}3\% & \cellcolor{typeblue!35}6\% & \cellcolor{targetteal!10}13\% & \cellcolor{targetteal!55}57\% & \cellcolor{targetteal!10}1\% & \cellcolor{targetteal!10}30\% \\
\textit{General/Ask} & 27 & 345 & 14.3\% & \cellcolor{typeblue!35}64\% & \cellcolor{typeblue!55}19\% & \cellcolor{typeblue!10}3\% & \cellcolor{typeblue!20}5\% & \cellcolor{typeblue!20}2\% & \cellcolor{typeblue!20}1\% & \cellcolor{typeblue!35}6\% & \cellcolor{targetteal!35}37\% & \cellcolor{targetteal!20}25\% & \cellcolor{targetteal!55}6\% & \cellcolor{targetteal!20}33\% \\
\textit{Misc} & 140 & 211 & 8.1\% & \cellcolor{typeblue!20}57\% & \cellcolor{typeblue!55}14\% & \cellcolor{typeblue!20}6\% & \cellcolor{typeblue!35}10\% & \cellcolor{typeblue!35}3\% & \cellcolor{typeblue!10}1\% & \cellcolor{typeblue!55}9\% & \cellcolor{targetteal!20}26\% & \cellcolor{targetteal!35}28\% & \cellcolor{targetteal!35}2\% & \cellcolor{targetteal!55}43\% \\
\textit{Humor} & 19 & 195 & 30.2\% & \cellcolor{typeblue!10}37\% & \cellcolor{typeblue!35}11\% & \cellcolor{typeblue!35}10\% & \cellcolor{typeblue!55}30\% & \cellcolor{typeblue!55}5\% & \cellcolor{typeblue!35}2\% & \cellcolor{typeblue!20}5\% & \cellcolor{targetteal!35}34\% & \cellcolor{targetteal!35}31\% & \cellcolor{targetteal!55}4\% & \cellcolor{targetteal!20}31\% \\
\textit{Geography} & 79 & 110 & 5.9\% & \cellcolor{typeblue!35}67\% & \cellcolor{typeblue!10}1\% & \cellcolor{typeblue!35}14\% & \cellcolor{typeblue!10}1\% & \cellcolor{typeblue!20}3\% & \cellcolor{typeblue!55}12\% & \cellcolor{typeblue!10}3\% & \cellcolor{targetteal!20}20\% & \cellcolor{targetteal!10}12\% & \cellcolor{targetteal!35}3\% & \cellcolor{targetteal!55}65\% \\
\textit{Hobbies} & 42 & 75 & 6.9\% & \cellcolor{typeblue!35}63\% & \cellcolor{typeblue!20}8\% & \cellcolor{typeblue!20}5\% & \cellcolor{typeblue!55}11\% & \cellcolor{typeblue!10}1\% & \cellcolor{typeblue!35}3\% & \cellcolor{typeblue!55}9\% & \cellcolor{targetteal!20}28\% & \cellcolor{targetteal!55}32\% & \cellcolor{targetteal!20}1\% & \cellcolor{targetteal!35}39\% \\
\textit{Politics} & 27 & 60 & 6.2\% & \cellcolor{typeblue!20}55\% & \cellcolor{typeblue!10}3\% & \cellcolor{typeblue!55}20\% & \cellcolor{typeblue!35}7\% & \cellcolor{typeblue!35}3\% & \cellcolor{typeblue!55}5\% & \cellcolor{typeblue!35}7\% & \cellcolor{targetteal!10}15\% & \cellcolor{targetteal!55}45\% & \cellcolor{targetteal!20}2\% & \cellcolor{targetteal!35}38\% \\
\textit{Beauty/Fashion} & 8 & 16 & 4.5\% & \cellcolor{typeblue!10}50\% & \cellcolor{typeblue!55}50\% & 0\% & 0\% & 0\% & 0\% & 0\% & \cellcolor{targetteal!55}56\% & \cellcolor{targetteal!20}25\% & 0\% & \cellcolor{targetteal!10}19\% \\
\midrule
\textbf{Total} & \textbf{615} & \textbf{3600} & \textbf{14.1\%} & 70\% & 9\% & 7\% & 6\% & 3\% & 2\% & 4\% & 33\% & 24\% & 2\% & 41\% \\
\bottomrule
\end{tabular}}
\caption{LLM-annotated posts that contain a stereotype, by community domain, stereotype type, and target racial group, ordered by stereotype count. \textbf{\#Stereo.}: number of posts labeled as containing a stereotype; \textbf{Stereo.\%}: share of the domain's posts that contain a stereotype.  \emph{Cult.}=cultural, \emph{App.}=appearance, \emph{Crim.}=criminality, \emph{Deh.}=dehumanization, \emph{Int.}=intelligence, \emph{Eco.}=economic, \emph{Oth.}=other. Rows may not sum to exactly 100\% due to rounding.} 
\label{tab:stereotype_rollup_v2}
\end{table*}

\begin{table}[t]
\centering
\small
\setlength{\tabcolsep}{4pt}
\renewcommand{\arraystretch}{1.1}
\begin{tabular}{lrrrrrr}
\toprule
\textbf{Gender} & \textbf{n} & \textbf{Auto} & \textbf{Hete} & \textbf{Homo} & \textbf{Hypo} & \textbf{Meta} \\
\midrule
Woman       & 208 & 87.0 & 0.5 & 2.9 & 0.0 & 9.6 \\
Man         & 199 & 82.9 & 1.0 & 6.5 & 0.0 & 9.5 \\
Undisclosed & 905 & 49.3 & 19.7 & 16.1 & 4.9 & 9.3 \\
\bottomrule
\end{tabular}
\caption{Author gender $\times$ narrative position among the 1{,}312
counter-stories. Percentages are row percentages: e.g., 87\% of
counter-stories written by women use autodiegetic narration. Gender is disclosed in 31\% of counter-stories ($n{=}407$).}
\label{tab:gender_position}
\end{table}

\begin{table}[t]
\centering
\small

\begin{tabular}{lcccc}
\toprule
 & \multicolumn{2}{c}{\textbf{Posts}} & \multicolumn{2}{c}{\textbf{Comments}} \\
\cmidrule(lr){2-3}\cmidrule(lr){4-5}
\textbf{Backbone} & Agr.\% & $\kappa$ & Agr.\% & $\kappa$ \\
\midrule
Gemma-3-27B    & 69 & 0.36 & 60 & 0.27 \\
OLMo-2-32B     & 86 & 0.73 & 68 & 0.41 \\
Qwen3-32B      & 85 & 0.69 & 77 & 0.56 \\
\rowcolor{gray!18}
\textbf{Claude Sonnet 4.6} & \textbf{88} & \textbf{0.76} & \textbf{86} & \textbf{0.72} \\
\bottomrule
\end{tabular}
\caption{LLM backbone comparison. Table shows agreement with human annotations on the validation set (50 posts, 100 comments). Claude Sonnet~4.6
(our pipeline backbone) shows the strongest agreement on both tasks.}
\label{tab:backbone_comparison}
\end{table}

\begin{table}[t]
\centering\small
\setlength{\tabcolsep}{6pt}\renewcommand{\arraystretch}{1.15}
\begin{tabular}{lcccc}
\toprule
\textbf{Task} & \textbf{P} & \textbf{R} & \textbf{F1} & \textbf{Acc} \\
\midrule
Stereotype    & 0.93 & 0.84 & 0.89 & 0.88 \\
Counter-story & 0.81 & 0.88 & 0.84 & 0.86 \\
\bottomrule
\end{tabular}
\caption{Precision, recall, F1, and accuracy of Claude Sonnet~4.6 against human-majority labels on the validation set.}
\label{tab:prf}
\end{table}

\begin{table}[t]
\centering\small
\setlength{\tabcolsep}{5pt}\renewcommand{\arraystretch}{1.15}
\begin{tabular}{lcccc}
\toprule
\textbf{Certainty} & \textbf{Ster.\,\%} & \textbf{Ster.\,Acc.} & \textbf{CS\,\%} & \textbf{CS\,Acc.} \\
\midrule
1--6  & 7  & 75\% & 6  & 33\% \\
7--8  & 24 & 79\% & 30 & 77\% \\
9--10 & 69 & 93\% & 64 & 95\% \\
\bottomrule
\end{tabular}
\caption{Accuracy by Sonnet's self-reported certainty on the validation set. Columns show the share of items in each certainty bin and the accuracy within it. Point-biserial correlation between certainty and correctness: $r{=}0.25$ (stereotype), $r{=}0.41$ (counter-story).}
\label{tab:calibration}
\end{table}

\begin{figure}[t]
\centering
\includegraphics[width=0.9\columnwidth]{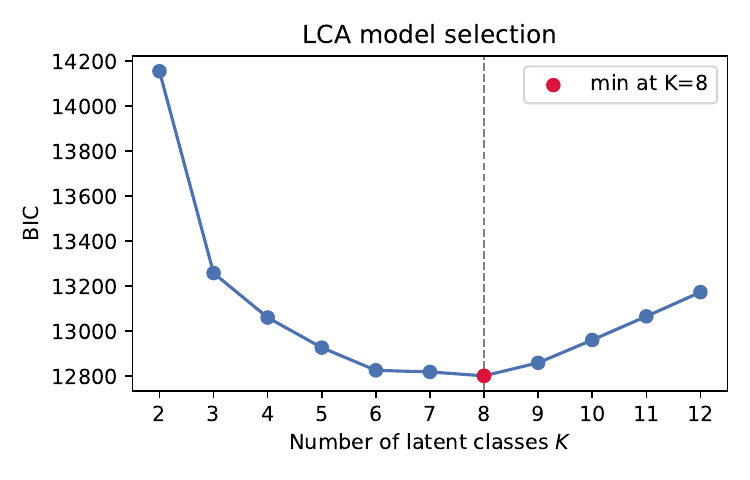}
\caption{BIC across $K{=}2$--$12$ for the LCA over the taxonomy features; BIC is minimized at $K{=}8$. We report seven roles after merging two adjacent classes.}
\label{fig:bic}
\end{figure}

\begin{table}[t]
\centering\small
\setlength{\tabcolsep}{5pt}\renewcommand{\arraystretch}{1.1}
\begin{tabular}{lrrr}
\toprule
\textbf{Subreddit} & \textbf{\#Story} & \textbf{\#CS} & \textbf{CS\%} \\
\midrule
\textbf{r/changemyview} & \textbf{183} & \textbf{129} & \textbf{70.5} \\
r/unpopularopinion      & 274 & 62  & 22.6 \\
r/politicalhumor        &  10 &  3  & 30.0 \\
r/conspiracy            &  42 &  2  &  4.8 \\
r/tiadiscussion         &   5 &  2  & 40.0 \\
r/negareddit            &   2 &  1  & 50.0 \\
r/shitredditsays        &  12 &  1  &  8.3 \\
r/badpeoplestories      &  11 &  0  &  0.0 \\
r/trueunpopularopinion  &  10 &  0  &  0.0 \\
r/subredditdrama        &   1 &  0  &  0.0 \\
\midrule
\textbf{Total} & \textbf{550} & \textbf{200} & \textbf{36.4} \\
\bottomrule
\end{tabular}
\caption{Counter-story rate (CS\%) by subreddit within the Debate/Opinion domain.
\#Story = story-containing comments; \#CS = counter-stories.}
\label{tab:cmv_breakdown}
\end{table}

\begin{table}[t]
\centering
\small
\setlength{\tabcolsep}{4pt}
\renewcommand{\arraystretch}{1.1}
\begin{tabular}{lrrrrrr}
\toprule
\textbf{Race} & \textbf{n} & \textbf{Auto} & \textbf{Hete} & \textbf{Homo} & \textbf{Hypo} & \textbf{Meta} \\
\midrule
Asian       & 371 & 77.1 & 3.8 & 8.1  & 0.3 & 10.8 \\
White       & 162 & 72.2 & 3.7 & 11.1 & 0.6 & 12.3 \\
Black       & 95  & 80.0 & 3.2 & 5.3  & 3.2 & 8.4  \\
Hispanic    & 52  & 69.2 & 3.8 & 7.7  & 0.0 & 19.2 \\
Other       & 45  & 71.1 & 4.4 & 11.1 & 0.0 & 13.3 \\
Undisclosed & 587 & 41.7 & 26.2 & 17.5 & 6.6 & 6.6 \\
\bottomrule
\end{tabular}
\caption{Author race $\times$ narrative position among the 1{,}312 counter-stories (row percentages). Race is disclosed in 55\% of counter-stories ($n{=}725$).}
\label{tab:race_position}
\end{table}

\begin{figure*}[t]
\centering
\includegraphics[width=\textwidth]{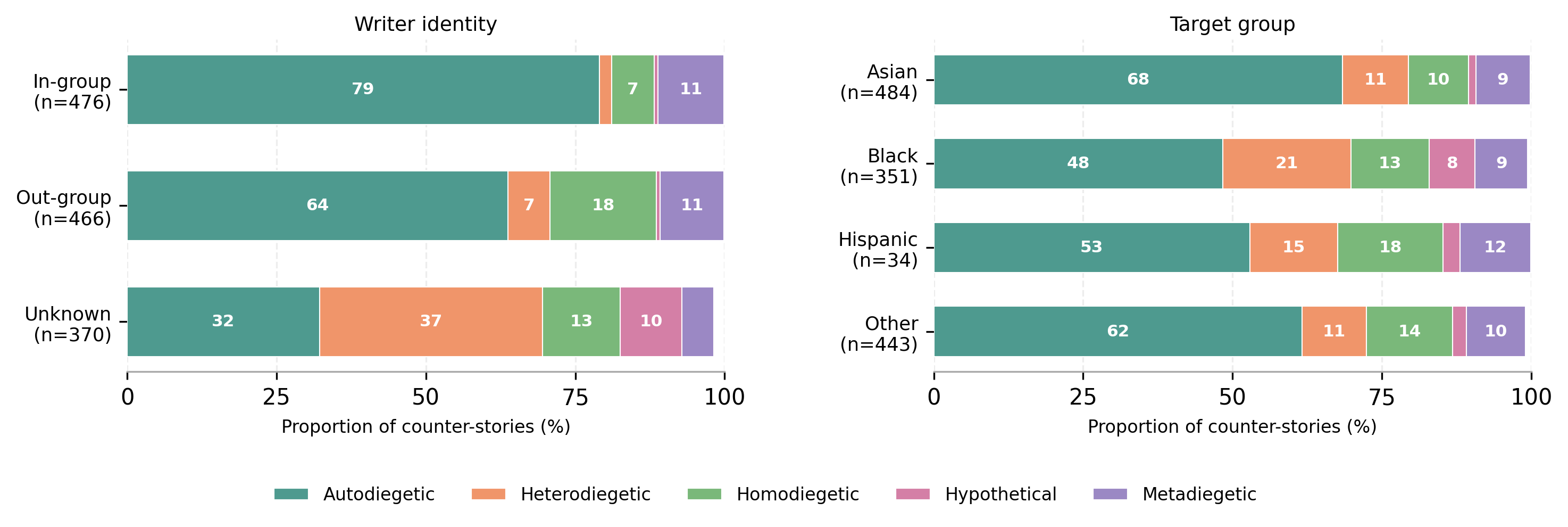}
\caption{Narrative position by writer identity (left) and target group (right), as a proportion of counter-stories within each group.}
\label{fig:position_breakdown}
\end{figure*}

\begin{table}[t]
\centering
\footnotesize
\setlength{\tabcolsep}{4pt}
\renewcommand{\arraystretch}{1.15}
\begin{tabularx}{\columnwidth}{@{}p{2.6cm}X@{}}
\toprule
\textbf{Category (\#subs)} & \textbf{Example subreddits} \\
\midrule
Misc (140)              & r/tipofmytongue, r/mildlyinfuriating, r/news, r/worldnews, r/mildlyinteresting \\
\addlinespace
Culture/Identity (139)  & r/asianparentstories, r/india, r/indiaspeaks, r/actuallesbians, r/mexico \\
\addlinespace
Advice/Edu/Work (91) & r/depression, r/tooafraidtoask, r/legaladvice, r/dating, r/relationship\_advice \\
\addlinespace
Geography (79)          & r/asknyc, r/houston, r/sandiego, r/casualuk, r/chicago \\
\addlinespace
Hobbies (42)            & r/squaredcircle, r/nba, r/cfb, r/trees, r/tattoos \\
\addlinespace
Stories (28)            & r/amitheasshole, r/trueoffmychest, r/fatpeoplestories, r/offmychest, r/vent \\
\addlinespace
General/Ask (27)        & r/askredditafterdark, r/nostupidquestions, r/askreddit, r/askmen, r/teenagers \\
\addlinespace
Politics (27)           & r/conservative, r/politics, r/wayofthebern, r/politicaldiscussion, r/thedeprogram \\
\addlinespace
Humor (19)              & r/jokes, r/showerthoughts, r/facepalm, r/asiansgonewild, r/shitamericanssay \\
\addlinespace
Debate/Opinion (15)     & r/conspiracy, r/unpopularopinion, r/changemyview, r/shitredditsays, r/politicalhumor \\
\addlinespace
Beauty/Fashion (8)      & r/skincareaddiction, r/curlyhair, r/femalefashionadvice, r/beautyguruchatter, r/asianbeauty \\
\bottomrule
\end{tabularx}
\caption{Community categories and example member subreddits, ordered by number of member subreddits. We show five most frequent subreddits per category.}
\label{tab:subreddit_categories}
\end{table}

\begin{table*}[t]
\centering
\small
\setlength{\tabcolsep}{4pt}
\renewcommand{\arraystretch}{1.1}
\begin{tabular}{lrrrrrrrr}
\toprule
\textbf{Domain} & \textbf{\#CS} & \makecell[c]{\textbf{Hypo-}\\\textbf{thetical}} & \makecell[c]{\textbf{Meta-}\\\textbf{diegetic}} & \makecell[c]{\textbf{Second-}\\\textbf{hand}} & \textbf{Exemplar} & \textbf{Fictional} & \makecell[c]{\textbf{Connection to}\\\textbf{Target Group}} & \makecell[c]{\textbf{Membership}\\\textbf{Undisclosed}}\\
\midrule
\textit{Culture/Identity} & 430 & \cellcolor{narrviolet!10}1\% & \cellcolor{narrviolet!25}10\% & \cellcolor{narrviolet!25}10\% & \cellcolor{narrviolet!10}4\% & \cellcolor{narrviolet!10}1\% & \cellcolor{identblue!25}12\% & \cellcolor{identblue!10}31\% \\
\textit{Advice/Education/Work} & 300 & \cellcolor{narrviolet!25}2\% & \cellcolor{narrviolet!10}6\% & \cellcolor{narrviolet!25}8\% & \cellcolor{narrviolet!10}3\% & \cellcolor{narrviolet!10}1\% & \cellcolor{identblue!80}29\% & \cellcolor{identblue!25}56\% \\
\textit{Debate/Opinion} & 200 & \cellcolor{narrviolet!50}8\% & \cellcolor{narrviolet!80}13\% & \cellcolor{narrviolet!25}10\% & \cellcolor{narrviolet!80}28\% & \cellcolor{narrviolet!50}4\% & \cellcolor{identblue!25}20\% & \cellcolor{identblue!50}76\% \\
\textit{General/Ask} & 140 & \cellcolor{narrviolet!10}1\% & \cellcolor{narrviolet!50}11\% & \cellcolor{narrviolet!80}14\% & \cellcolor{narrviolet!25}6\% & \cellcolor{narrviolet!10}1\% & \cellcolor{identblue!50}26\% & \cellcolor{identblue!25}50\% \\
\textit{Stories} & 93 & \cellcolor{narrviolet!25}5\% & \cellcolor{narrviolet!50}11\% & \cellcolor{narrviolet!50}12\% & \cellcolor{narrviolet!25}6\% & 0\% & \cellcolor{identblue!50}23\% & \cellcolor{identblue!25}54\% \\
\textit{Geography} & 53 & \cellcolor{narrviolet!50}8\% & \cellcolor{narrviolet!50}11\% & \cellcolor{narrviolet!50}13\% & \cellcolor{narrviolet!50}11\% & 0\% & \cellcolor{identblue!50}21\% & \cellcolor{identblue!50}74\% \\
\textit{Humor} & 39 & \cellcolor{narrviolet!25}5\% & \cellcolor{narrviolet!10}5\% & \cellcolor{narrviolet!50}13\% & \cellcolor{narrviolet!50}15\% & \cellcolor{narrviolet!50}10\% & \cellcolor{identblue!80}31\% & \cellcolor{identblue!80}92\% \\
\textit{Hobbies} & 24 & 0\% & 0\% & \cellcolor{narrviolet!10}4\% & 0\% & \cellcolor{narrviolet!80}12\% & \cellcolor{identblue!10}4\% & \cellcolor{identblue!80}92\% \\
\textit{Misc} & 15 & 0\% & \cellcolor{narrviolet!25}7\% & \cellcolor{narrviolet!10}7\% & \cellcolor{narrviolet!25}7\% & 0\% & \cellcolor{identblue!25}13\% & \cellcolor{identblue!25}60\% \\
\textit{Politics} & 12 & 0\% & 0\% & 0\% & \cellcolor{narrviolet!80}42\% & 0\% & \cellcolor{identblue!10}8\% & \cellcolor{identblue!80}92\% \\
\textit{Beauty/Fashion} & 6 & \cellcolor{narrviolet!80}17\% & \cellcolor{narrviolet!80}17\% & 0\% & \cellcolor{narrviolet!50}17\% & 0\% & 0\% & \cellcolor{identblue!50}67\% \\
\midrule
\textbf{Total} & \textbf{1312} & 3\% & 9\% & 10\% & 9\% & 2\% & 20\% & 53\% \\
\bottomrule
\end{tabular}
\caption{Counter-story narrative positions and subtypes omitted from Table~\ref{tab:domain_rollup_combined} (row \%\ of \#CS), by community domain: hypothetical and metadiegetic narration; the second-hand, exemplar/celebrity, and fictional subtypes; the share of writers reporting a personal connection to the target group; and the share whose group identity is \emph{undisclosed} (neither target- nor dominant-group membership disclosed), complementing the Target/Dominant columns in Table~\ref{tab:domain_rollup_combined}.}
\label{tab:domain_rollup_appendix}
\end{table*}

\subsection{Counter-Story Regression}
\label{appendix:regression}
\begin{figure}[t]
\centering
\includegraphics[width=\columnwidth]{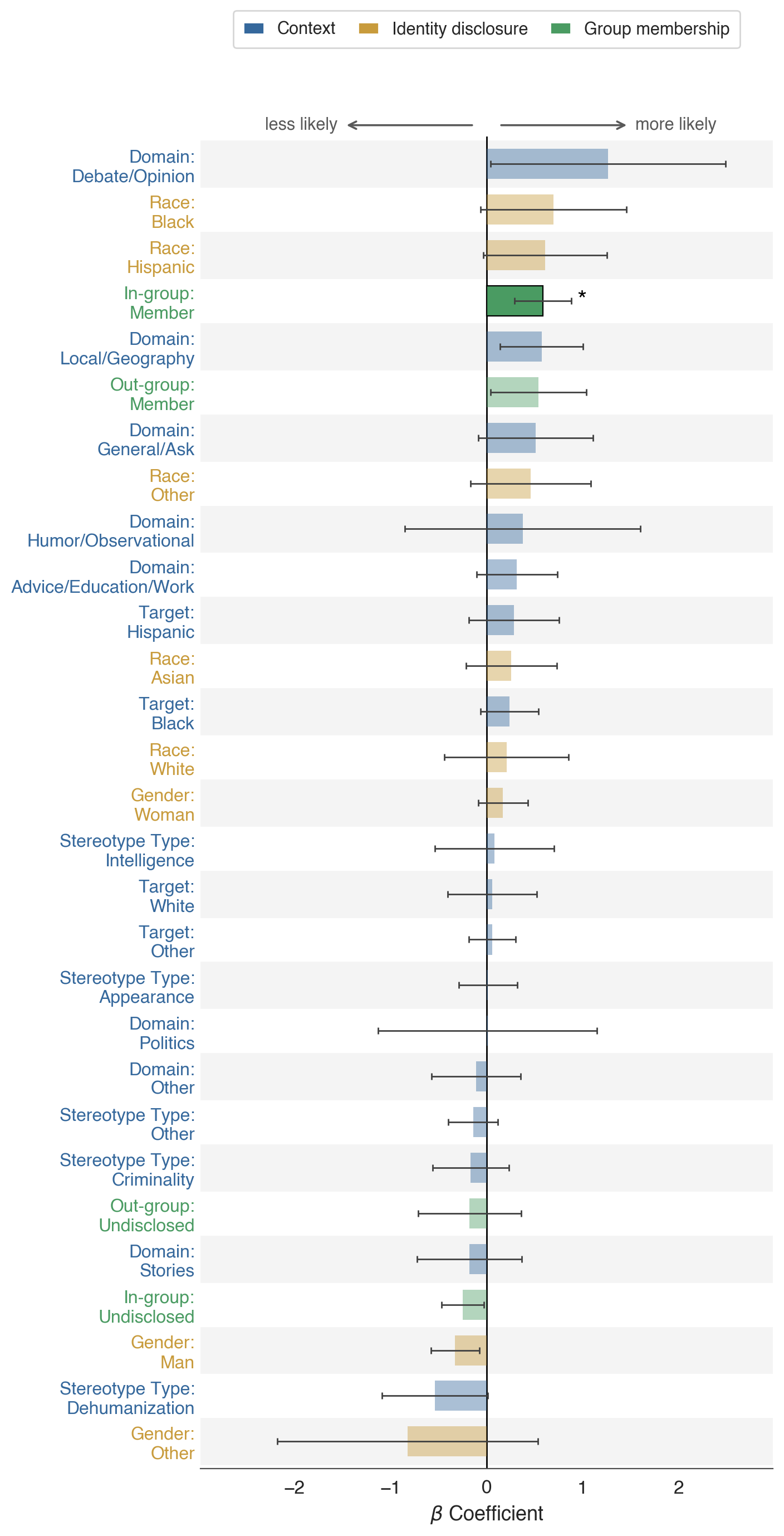}
\caption{Logistic-regression coefficients (log-odds) for the model using both context and speaker predictor predicting whether a story comment is counter-story or not. Bars with $\ast$ are significant after Holm--Bonferroni correction ($p<0.01$).}
\label{fig:forest_joint}
\end{figure}

\begin{figure}[t]
\centering
\includegraphics[width=\columnwidth]{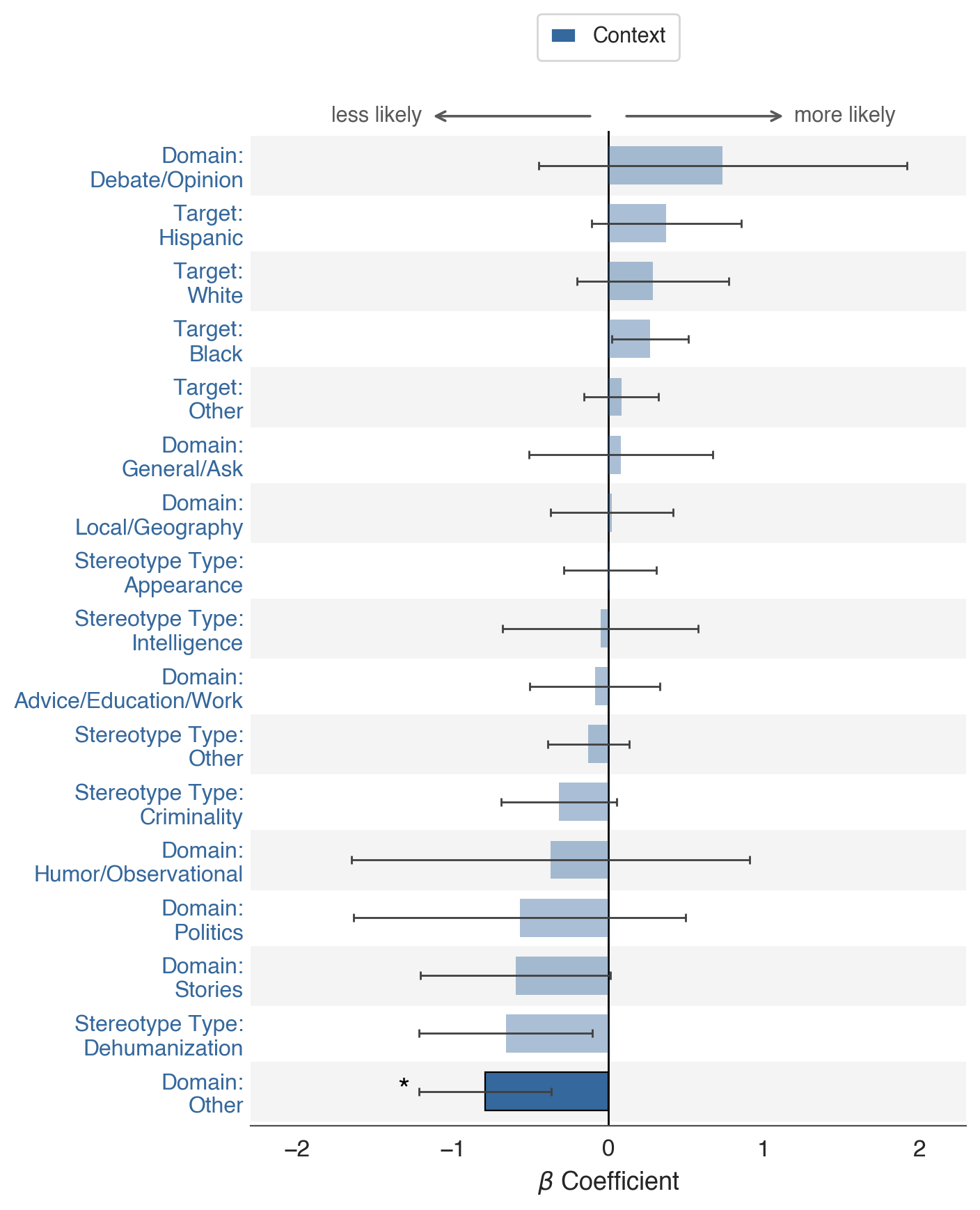}
\caption{Logistic-regression coefficients (log-odds) for the model using context predictors only to predict whether a story comment is counter-story or not. Bars with $\ast$ are significant after Holm--Bonferroni correction ($p<0.01$).}
\label{fig:forest_context}
\end{figure}

\begin{figure}[t]
\centering
\includegraphics[width=\columnwidth]{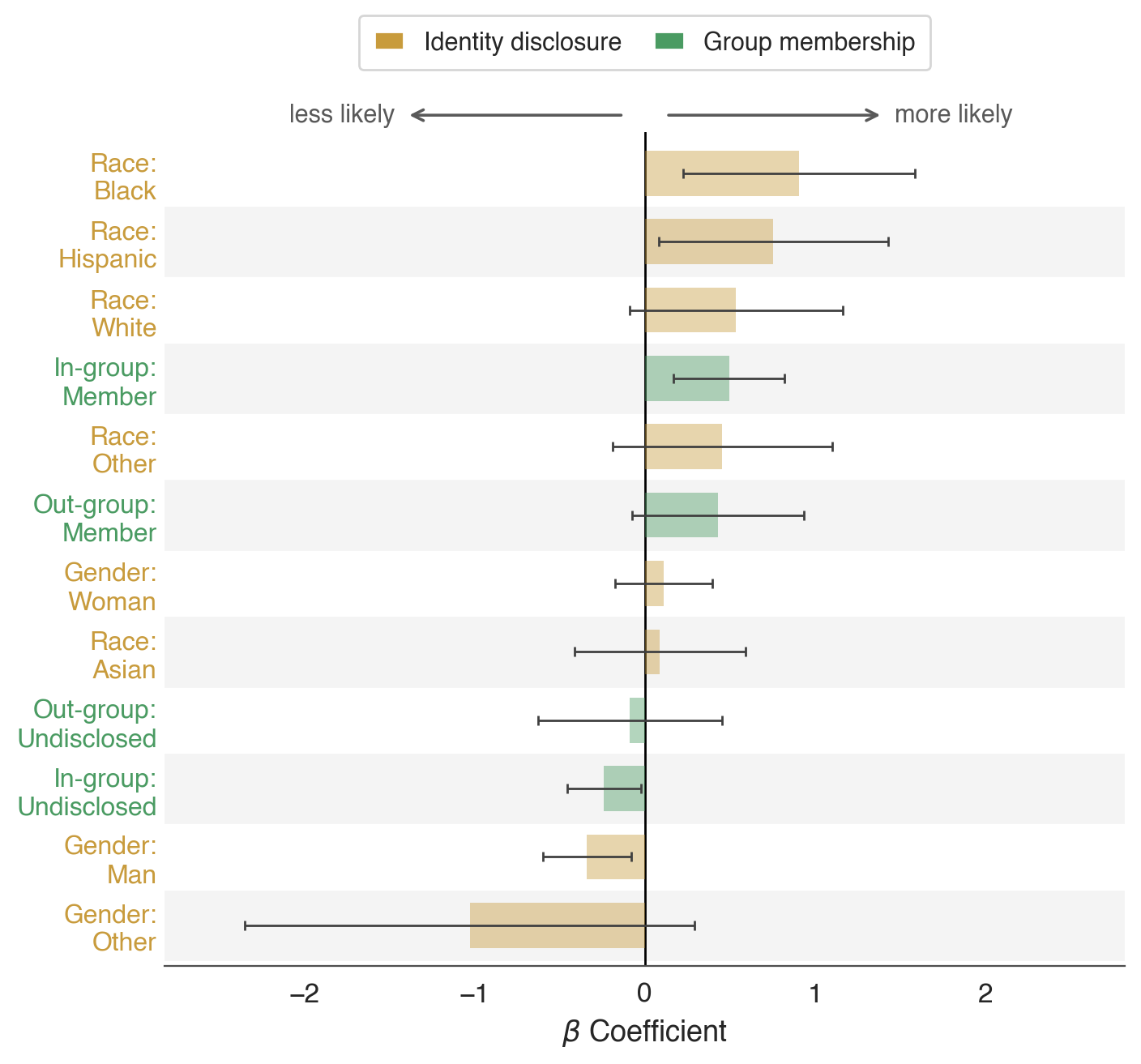}
\caption{Logistic-regression coefficients (log-odds) for the \emph{speaker-only} model predicting counter-story status. Bars with $\ast$ are significant after Holm--Bonferroni correction ($p<0.01$).}
\label{fig:forest_speaker}
\end{figure}

We fit logistic regressions predicting whether a story comment is a counter-story ($N{=}6{,}237$; $1{,}312$ counter-stories), with cluster-robust standard errors by subreddit and Holm--Bonferroni correction at $p<0.01$. We use \emph{context} predictors (community domain, stereotype type, target group), \emph{speaker} predictors (disclosed race, gender, in-/out-group membership), and a \emph{joint} model (using both types of predictors). Speaker predictors account for roughly twice the variance of context predictors (pseudo-$R^2$ $0.048$ vs.\ $0.026$; joint $0.074$), indicating that who is speaking matters more than where or what they respond to (Fig.~\ref{fig:forest_speaker}, ~\ref{fig:forest_context}). In-group membership is the only predictor that survives Holm--Bonferroni correction (OR\,$=1.80$, $p_{\mathrm{Holm}}<0.01$): writers who identify with the targeted group are more likely to write counter-stories (Fig.~\ref{fig:forest_joint}). This suggests that counter-storytelling is shaped by the interaction of speaker position, audience, stereotype type, and community context, rather than by any single dimension.

\begin{table}[t]
\centering\small
\begin{tabular}{lrrrr}
\toprule
\textbf{Stereotype Type} & \textbf{Asian} & \textbf{Black} & \textbf{Hispanic} & \textbf{Other} \\
\midrule
Cultural       & 74.3\% & 52.6\% & 73.2\% & 76.5\% \\
Appearance     & 10.6\% & 12.4\% &  5.6\% &  5.7\% \\
Criminality    &  2.9\% & 13.6\% & 11.3\% &  6.6\% \\
Dehumanization &  4.7\% &  9.4\% &  5.6\% &  3.9\% \\
Other          &  3.4\% &  6.8\% &  2.8\% &  2.9\% \\
Intelligence   &  3.2\% &  3.1\% &  0.0\% &  1.8\% \\
Economic       &  0.9\% &  2.0\% &  1.4\% &  2.6\% \\
\midrule
$n$ posts & 1{,}198 & 868 & 71 & 1{,}463 \\
\bottomrule
\end{tabular}
\caption{Full stereotype-type distribution across all race-groups.}
\label{tab:stereotype_dist}
\end{table}

\clearpage

\end{document}

%% file: commands.tex
\definecolor{maria-color}{HTML}{7881F2}

\usepackage{xspace}
\definecolor{todo-color}{HTML}{ff3232}

\usepackage{xparse}
\NewDocumentCommand{\todocite}{g}{%
  {\color{todo-color}[CITE\IfNoValueF{#1}{: #1}]}\xspace%
}

\definecolor{lightblue}{RGB}{212, 235, 255}
\definecolor{orange}{RGB}{255, 105, 0}
\definecolor{lightgreen}{RGB}{177, 231, 171}
\definecolor{lightyellow}{RGB}{255, 255, 148}

\usepackage{booktabs}
\usepackage{array}

\usepackage{float}
\usepackage{subcaption}
\usepackage{graphicx}

\definecolor{applications-color}{HTML}{e6d4fa}
\definecolor{theory-color}{HTML}{cff5f5}
\definecolor{proposal-color}{HTML}{d4fae6}
\definecolor{4-color}{HTML}{f9d4fa}
\definecolor{5-color}{HTML}{fae6d4}
\definecolor{warn-color}{HTML}{f9d4fa}

\usepackage{enumitem}
\usepackage{tcolorbox}